\documentclass[lettersize,journal]{IEEEtran}
\usepackage{amsmath,amsfonts}
\usepackage{algorithmic}
\usepackage{algorithm}
\usepackage{array}
\usepackage[caption=false,font=normalsize,labelfont=sf,textfont=sf]{subfig}
\usepackage{textcomp}
\usepackage{stfloats}
\usepackage{url}
\usepackage{verbatim}
\usepackage{graphicx}
\usepackage{cite}
\usepackage{amsmath, amsfonts, epsfig}
\usepackage{amssymb}
\usepackage{xcolor}
\usepackage{multirow}
\usepackage{graphicx}
\usepackage{amsthm}
\usepackage{bm}
\usepackage{algorithmic}
\usepackage{algorithm}
\usepackage{array}
\usepackage{subfig}
\usepackage{textcomp}
\usepackage{tabularx}
\usepackage{stfloats}
\usepackage{pgfplots}
\usepackage{url}
\usepackage{cite}
\usepackage[colorlinks,urlcolor=blue]{hyperref}

\definecolor{lime}{HTML}{A6CE39}
\DeclareRobustCommand{\orcidicon}{
	\begin{tikzpicture}
		\draw[lime, fill=lime] (0,0)
		circle[radius=0.16]
		node[white]{{\fontfamily{qag}\selectfont \tiny \.{I}D}};
	\end{tikzpicture}
	\hspace{-2mm}
}
\foreach \x in {A, ..., Z}{%
	\expandafter\xdef\csname orcid\x\endcsname{\noexpand\href{https://orcid.org/\csname orcidauthor\x\endcsname}{\noexpand\orcidicon}}
}

\begin{document}
\title{Deep Robust Reversible Watermarking}

\author{Jiale Chen,~~Wei Wang,~~\IEEEmembership{Member,~IEEE,} Chongyang Shi,~~\IEEEmembership{Member,~IEEE,}
\\
Li Dong,~~\IEEEmembership{Member,~IEEE,} Yuanman Li,~~\IEEEmembership{Senior Member,~IEEE,} Xiping Hu,~~\IEEEmembership{Member,~IEEE}

\thanks{Jiale Chen is with the School of Computer Science, Beijing Institute of Technology, Beijing 100081, China, and also with the Guangdong-Hong Kong-Macao Joint Laboratory for Emotional Intelligence and Pervasive Computing, Artificial Intelligence Research Institute, Shenzhen MSU-BIT University, Shenzhen 518172, China (e-mail: chenoly@outlook.com).}

\thanks{Wei Wang and Xiping Hu are with the School of Medical Technology, Beijing Institute of Technology, Beijing 100081, China, and also with the Guangdong-Hong Kong-Macao Joint Laboratory for Emotional Intelligence and Pervasive Computing, Artificial Intelligence Research Institute, Shenzhen MSU-BIT University, Shenzhen 518172, China (e-mail: ehomewang@ieee.org; huxp@bit.edu.cn). }

\thanks{Chongyang Shi is with the School of Computer Science, Beijing Institute of Technology, Beijing 100081, China (e-mail: cy\_shi@bit.edu.cn).}

\thanks{Yuanman Li is with the College of Electronics and Information Engineering, Shenzhen
University, Shenzhen 518060, China. (email: yuanmanli@szu.edu.cn).}
\thanks{Li Dong is with the Department of Computer Science, Faculty of Electrical Engineering and Computer Science, Ningbo University, Ningbo 315211, China (e-mail: dongli@nbu.edu.cn).}
}

\maketitle

\begin{abstract}
Robust Reversible Watermarking (RRW) enables perfect recovery of cover images and watermarks in lossless channels while ensuring robust watermark extraction under lossy channels. However, existing RRW methods, mostly non-deep learning-based, suffer from complex designs, high computational costs, and poor robustness limiting their practical applications. To address these issues, this paper proposes Deep Robust Reversible Watermarking (DRRW), a deep learning-based RRW scheme. DRRW introduces an Integer Invertible Watermark Network (iIWN) to achieve an invertible mapping between integer data distributions, fundamentally addressing the limitations of conventional RRW approaches.
Unlike traditional RRW methods requiring task-specific designs for different distortions, DRRW adopts an encoder-noise layer-decoder framework, enabling adaptive robustness against various distortions through end-to-end training. During inference, the cover image and watermark are mapped into an overflowed stego image and latent variables. Arithmetic coding efficiently compresses these into a compact bitstream, which is embedded via reversible data hiding to ensure lossless recovery of both the image and watermark.
To reduce pixel overflow, we introduce an overflow penalty loss, significantly shortening the auxiliary bitstream while improving both robustness and stego image quality. Additionally, we propose an adaptive weight adjustment strategy that eliminates the need to manually preset the watermark loss weight, ensuring improved training stability and performance.
Experiments on multiple datasets demonstrate that DRRW achieves notable performance advantages. Compared to state-of-the-art RRW methods, DRRW improves robustness and reduces embedding, extraction, and recovery complexities by 55.14\(\times\), 5.95\(\times\), and 3.57\(\times\), respectively. The auxiliary bitstream is shortened by  43.86\(\times\), and reversible embedding succeeds on 16{,}762 images in the PASCAL VOC 2012 dataset, marking a significant step toward practical RRW applications. Notably, DRRW achieves robustness and visual quality superior to existing irreversible robust watermarking methods while strictly maintaining reversibility. The source code is available at \href{https://github.com/chenoly/Deep-Robust-Reversible-Watermark}{\textcolor{blue}{https://github.com/chenoly/Deep-Robust-Reversible-Watermark}}.
\end{abstract}

\vspace{-10pt}

\section{Introduction}
 \begin{figure}[t]
	\centering
	\includegraphics[width=\linewidth]{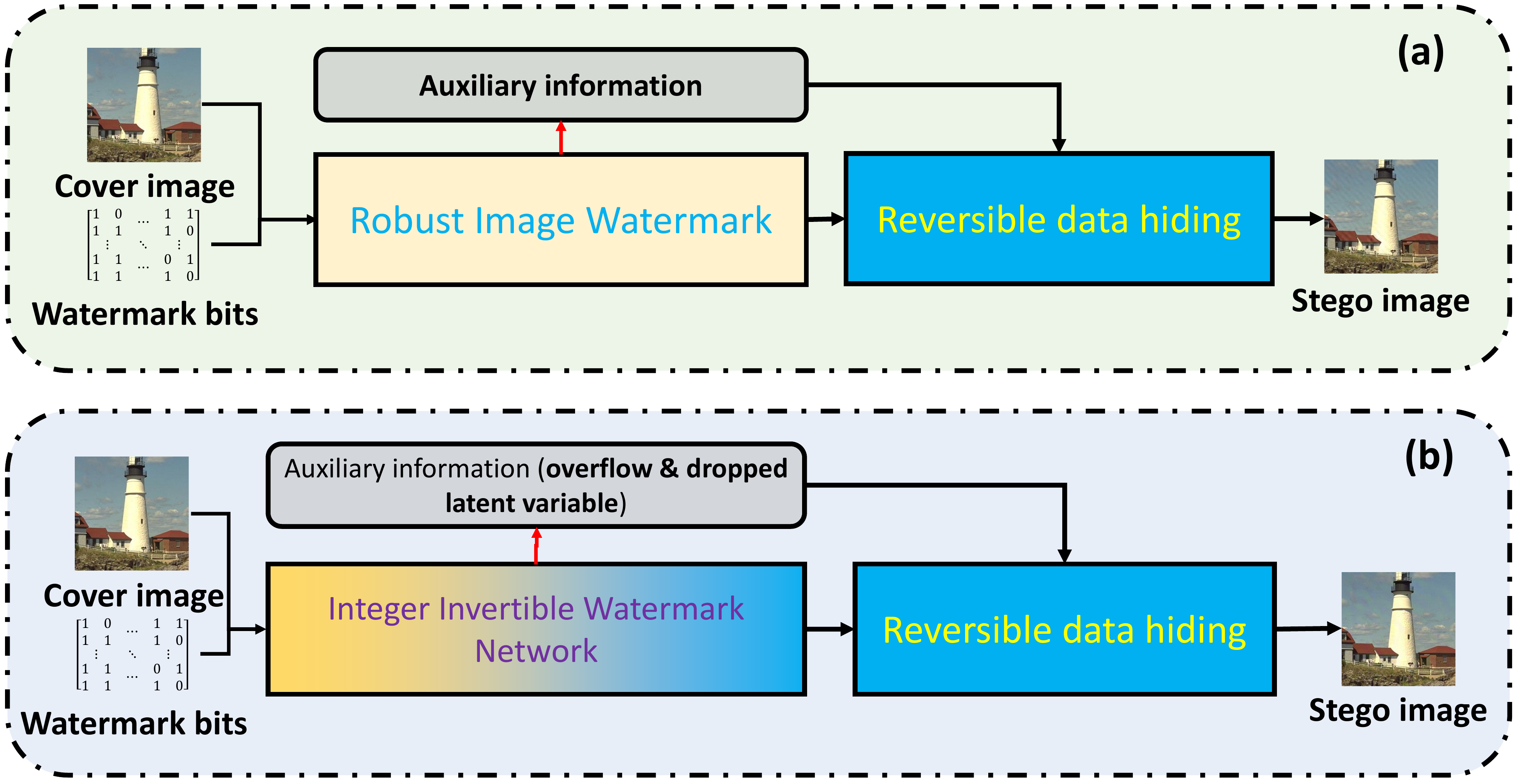}
    \caption{Comparison between traditional RRW and the proposed DRRW. (a) The traditional method embeds a robust watermark using conventional robust image watermarking methods; (b) The proposed DRRW embeds a robust watermark using an integer invertible watermark network.}
	\label{fig:intro_rrw_compare}
\end{figure}
\IEEEPARstart{W}{ith} the rapid evolution of digital content and internet technologies, safeguarding intellectual property and ensuring data security have become pressing concerns. Digital watermarking has emerged as a pivotal solution, embedding identifiable markers directly into multimedia content. This technique facilitates ownership verification, content integrity authentication, and secure dissemination of digital media \cite{podilchuk2001digital, singh2013survey}.
Robust Reversible Watermarking (RRW) has gained prominence due to its unique ability to maintain robustness against distortions while ensuring the lossless recovery of the original cover and watermark bits. This dual functionality makes RRW indispensable for high-stakes applications such as data integrity and copyright verification are critical, such as secure content transmission, digital media storage, and scenarios requiring absolute fidelity, such as medical image processing \cite{menendez2019survey,yan2022multiwatermarking}.

Traditional RRW methods can be broadly categorized into two directions: histogram shifting-based RRW \cite{liang2020robust,ni2008robust,de2003circular,gao2011lossless,an2012robust,hu2018new,zeng2010lossless} and two-stage embedding-based RRW \cite{hu2020cover,coltuc2007distortion,wang2019independent,kumar2020robust,tang2024robust}. Among them, the two-stage embedding strategy is the most representative. In this approach, the first stage embeds the watermark bits into the cover image using a robust watermarking technique, while the second stage conceals the auxiliary information required for cover image recovery into the stego image through reversible data hiding \cite{ni2006reversible} as illustrated in Figure \ref{fig:intro_rrw_compare}(a). Although these methods achieve a certain balance between robustness and reversibility, they often involve intricate designs and result in increased embedding time, limiting their practicality. 

\begin{figure}[t]
	\centering
	\includegraphics[width=\linewidth]{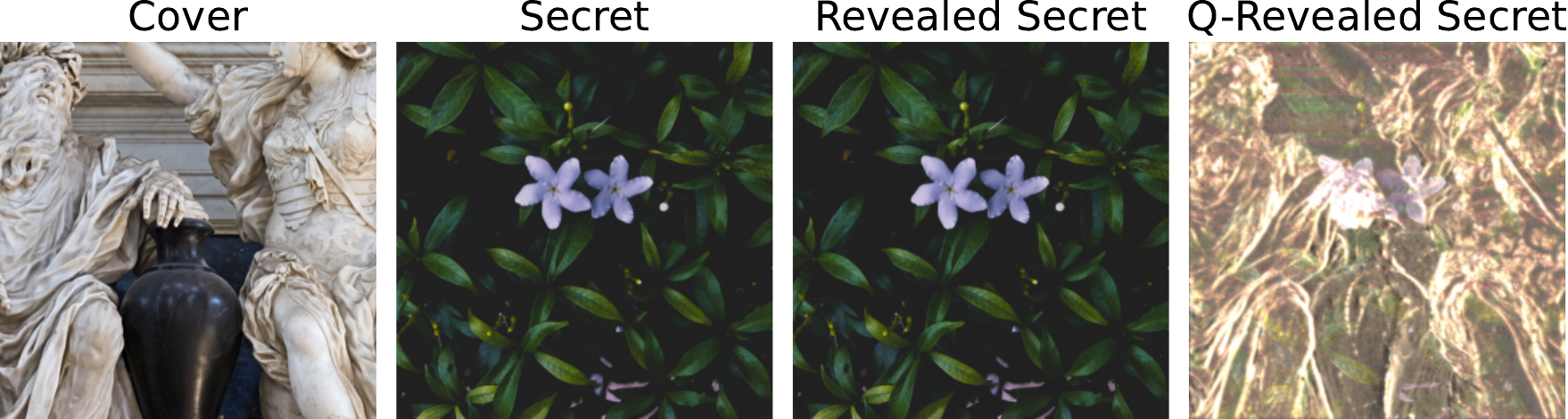}
    \caption{Comparison of \emph{revealed secret} (without stego quantization) and \emph{Q-revealed secret} (with stego quantization) in real-valued flow-based HiNet~\cite{jing2021hinet}.}
	\label{fig:intro_img_compare}
\end{figure}

Recently, Invertible Neural Networks (INNs) have gained widespread attention in fields such as digital watermarking \cite{fang2023flow,ma2022towards} and steganography \cite{jing2021hinet,lu2021large,xu2022robust,guan2022deepmih,li2024lidinet,li2023iscmis}. These networks serve as an invertible function to support invertible mappings between cover images-watermark pair and stego image, making them highly suitable for steganography and watermark tasks. INNs are particularly appealing because they can embed the watermark while maintaining the high imperceptibility of the original cover, which is crucial for applications in copyright protection and secure data transmission. Despite these advantages, a significant challenge arises from the majority of current INN-based image data hiding methods, which operate in the real-value domain, consequently, these methods also can be called real-valued flow-based data hiding. The bad thing is that these real-valued flow-based methods can only map real-valued data distributions, leading to an inherent issue in image storage: the need to quantize the generated stego images into 8-bit integers for storage. This quantization introduces information loss, which significantly impacts the performance of real-valued flow-based steganography. For instance, we evaluate the effects of quantization on the classic image-to-image steganography method, HiNet \cite{jing2021hinet}. As illustrated in Figure \ref{fig:intro_img_compare}, when the stego image is not restricted to the valid \(8\)-bit pixel value, the revealed secret image exhibits exceptional visual quality. However, after quantizing the stego image, the visual quality of the revealed secret image degrades significantly. We analyze the PSNR of the revealed secrets as shown in Figure \ref{fig:intro_psnr}. We can observe that the PSNR of the revealed secrets has a dramatic drop of about \(30\) dB after restricting the stego image to an 8-bit image format. This experiment highlights the profound impact of quantization on steganography, showing that secret information is often hidden within the fractional components of pixel values in stego images in INN-based steganography. Consequently, storing unquantized stego images in formats such as HDR or TIFF may alleviate this issue, albeit at the cost of reduced storage efficiency. To sum up, this quantization-induced information loss not only impairs the performance of INNs in image steganography but also restricts their applicability in robust reversible watermarking tasks.

To address this issue, we propose a novel two-stage robust reversible watermarking framework, termed Deep Robust Reversible Watermarking (DRRW), as illustrated in Figure \ref{fig:intro_rrw_compare}(b). The core of DRRW is the adoption of integer Invertible Watermark Networks (iIWNs), which enable lossless, invertible mappings between integer-discrete data distributions. By establishing an invertible mapping between the cover image-watermark pair and the stego image, iIWN fundamentally eliminates the irreversibility issues caused by quantization errors in real-valued flow-based data hiding methods. 
In training, DRRW employs an encoder-noise layer-decoder framework to learn robustness against various distortions. During inference, iIWN maps the cover image-watermark pair to an overflow stego image and a latent variable, and then three challenges need to be solved:  
1) Pixel values in the overflow stego image may exceed the valid range of $[0, 255]$;  
2) The latent variable must be losslessly stored within the stego image;  
3) The auxiliary information needs to be compressed into a short bitstream for embedding into the clipped stego image by RDH.  
To address these challenges, we introduce an overflow penalty loss during training to constrain pixel values within the allowable range. Additionally, for overflow pixels, we compute the overflow map and compress it and the latent variable into a compact bitstream by arithmetic coding \cite{witten1987arithmetic}, which is then embedded into the clipped stego image using RDH. This ensures lossless reconstruction of the cover image and embedded watermark.

The main contributions of this work are as follows:
 \begin{figure}[t]
	\centering
	\includegraphics[width=\linewidth]{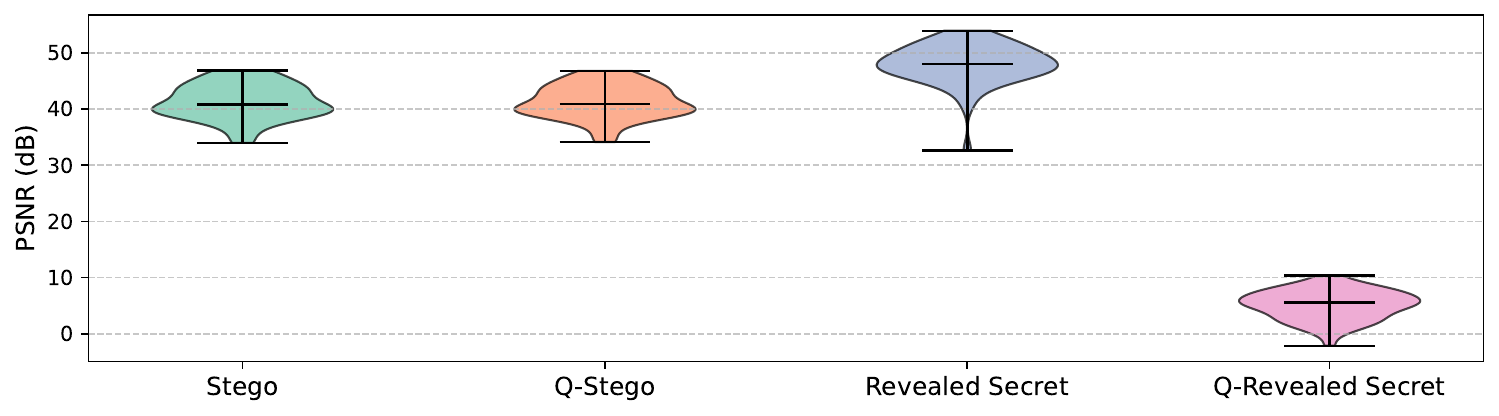}
    \caption{PSNR distributions of \emph{revealed secret} (without stego quantization) and \emph{Q-revealed secret} (with stego quantization) in real-valued flow-based HiNet~\cite{jing2021hinet}.}
	\label{fig:intro_psnr}
    \vspace{-15pt}
\end{figure}

\begin{itemize}
    \item We propose a two-stage robust reversible watermarking framework called Deep Robust Reversible Watermarking (DRRW). DRRW leverages an integer invertible watermark network to learn robustness under various distortions and achieve lossless and invertible mappings between the cover image-watermark pair and stego image.

    \item We introduce an overflow penalty loss during the training of DRRW. This loss not only significantly reduces the length of the auxiliary bitstream, but also improves the visual quality of stego images and enhances the robustness of the watermark, achieving a threefold benefit.

    \item We propose a training strategy with automatic adjustment of the watermark loss weight in DRRW. This strategy adaptively adjusts watermark loss weight based on the training process, eliminating the need for manual hyperparameter tuning and ensuring the model converges to optimal performance under different settings.

    \item Experimental results demonstrate the superior effectiveness of the DRRW framework, outperforming conventional robust reversible and irreversible watermarking methods. The framework has been successfully applied to large-scale datasets, proving its practical utility and scalability.
\end{itemize}

The rest of this paper is organized as follows: Section \ref{sec:related_work} reviews related work, Section \ref{sec:proposed_method} describes the DRRW framework, Section~\ref{sec:exp} presents experimental results, and Section~\ref{sec:conclusion} concludes the paper.

\section{Related Work}\label{sec:related_work}
\subsection{Robust Reversible Watermarking}
Reversible data hiding (RDH) is a form of watermarking designed to allow both the lossless recovery of the original cover and the extraction of embedded watermark bits in the lossless channel. Early contributions to RDH largely focused on spatial-domain techniques, exemplified by methods like Difference Expansion (DE) \cite{tian2003reversible}, Histogram Shifting (HS) \cite{ni2006reversible}, and Prediction-Error Expansion (PEE) \cite{sachnev2009reversible}. However, as applications began to demand robustness in lossy channels, robust reversible watermarking (RRW) emerged as a specialized extension of RDH. RRW not only ensures lossless recovery of the cover image and watermark in lossless channels but also enhances the resistance of the watermark to distortions in lossy channels. The dual goals of robustness and reversibility impose additional complexity compared to traditional robust watermarking approaches.

Mainstream RRW methods include histogram shifting-based RRW\cite{liang2020robust,ni2008robust,de2003circular,gao2011lossless,an2012robust,hu2018new,zeng2010lossless} and two-stage embedding-based RRW\cite{hu2020cover,coltuc2007distortion,wang2019independent,kumar2020robust,wang2024RobustRW,fu2023robust,tang2024robust,tang2022highly}. HS-based methods embed watermarks by manipulating robust image features through histogram shifting. Such as Pioneering studies in this domain, such as Ni \textit{et al.} introduced a robust lossless data hiding method using statistical quantities derived from patchwork theory, achieving both robustness to JPEG compression and losslessness\cite{ni2008robust}. Gao \textit{et al.} \cite{gao2011lossless} proposed a lossless data embedding framework leveraging generalized statistical quantity histograms, offering adaptability to diverse image types and scalability for varying capacities. The approach enhances robustness against JPEG compression. Furthermore, An \emph{et al.} \cite{an2012robust} introduced WSQH-SC, a robust reversible watermarking framework employing wavelet-domain histogram shifting and clustering. The method balances robustness and invisibility through enhanced pixel-wise masking, achieving superior performance across various image types.
These strategies excel in their simplicity and effectiveness but often struggle with scalability for complex distortions.

In contrast, two-stage embedding-based RRWs focus on integrating robust watermarking with reversible data hiding. This strategy was first proposed by Coltuc\cite{coltuc2007distortion}. Firstly the watermark is embedded by using a robust watermarking scheme and subsequently hides the bitstream required for recovery of cover and watermark into the stego image through RDH. Such as, Kumar \emph{et al.}\cite{kumar2020robust} introduced a RRW scheme leveraging a two-layer embedding strategy. By embedding secret data into the higher significant bit plane using PEE, the scheme ensures robustness against non-malicious attacks like JPEG compression while achieving high embedding capacity. Hu \emph{et al.} \cite{hu2020cover} proposed a cover-lossless robust watermarking scheme leveraging low-order Zernike moments to resist geometric deformations like scaling and rotation. Recently, Tang \emph{et al.}\cite{tang2024robust} proposed an RRW scheme that enhances watermark robustness and embedding capacity by using attack-simulation-based adaptive normalization. Their method utilizes polar harmonic transform moments as watermark cover and optimizes embedding strength through multi-level quantization in spread transform dither modulation. This method has robustness against common signal processing and geometric deformations.
However, these schemes are often high time-complexity and complex in design.

In addition, two-stage embedding-based RRWs using deep learning-based robust watermarking \cite{huang2022two,liu2023two} first embed watermarks via CNN-based robust watermarking. Due to the irreversibility of CNN-based processes, the residual image between the cover and stego image is then embedded into the stego image through RDH. However, these methods challenge RDH capacity, making them impractical for large-scale dataset applications.

\subsection{Real-Valued Flow-Based Data Hiding}
Normalizing flow has emerged as a powerful tool for both density estimation \cite{dinh2016density} and generative modeling \cite{kingma2018glow,dinh2015nice}. These models leverage invertible mapping constructed through coupling layers, enabling complex data distributions to be mapped into simple, tractable ones with precise likelihood evaluation \cite{dinh2015nice}. The dual-input structure of coupling layers makes flow-based models particularly effective in information hiding tasks, such as robust image watermarking \cite{fang2023flow,ma2022towards} and image-in-image steganography \cite{jing2021hinet,lu2021large,xu2022robust,guan2022deepmih,li2024lidinet,li2023iscmis}. However, existing flow-based watermarking and steganography methods primarily rely on real-valued computations throughout their forward and inverse processes. Consequently, although cover images are discrete and integer-valued, the outputs of these models are continuous-valued, necessitating a quantization step to map the outputs to valid pixel intensities for image storage and practical usage. This quantization process inevitably results in information loss, hindering the exact recovery of cover images in reversible watermarking applications. Recent studies have attempted to minimize the residual error between cover and stego images through real-valued flow-based approaches \cite{li2024lidinet}. Nevertheless, these methods still fail to entirely circumvent performance degradation stemming from the information loss induced by quantization, particularly in high-fidelity recovery scenarios where precision is critical.

\subsection{Integer Discrete Flows}
Integer Discrete Flows (IDFs) \cite{hoogeboom2019integer} are specifically tailored for modeling discrete data by employing integer-only operations in both forward and inverse transformations. This design ensures that the data remains in integer form throughout the entire encoding and decoding process, thereby eliminating the need for a quantization step and avoiding the information loss typically introduced in real-valued flow models. Unlike real-valued flows, which rely on continuous-valued coupling layers, IDFs utilize integer coupling layers to better preserve the integrity and precision of discrete data representations. This integer-based approach has garnered significant attention and recognition in fields such as lossless data compression \cite{ho2019compression,hoogeboom2019integer,berg2020idf++,zhang2021ivpf,zhang2021iflow} and generative modeling \cite{hoogeboom2021argmax}. In the context of reversible watermarking and information hiding, IDFs are particularly promising because their lack of quantization error and inherent reversibility align closely with the stringent requirements of such tasks, especially when exact cover image recovery and distortion-free reconstruction are crucial.

\section{Proposed Deep Robust Reversible Watermarking}\label{sec:proposed_method}  
\begin{figure}[t!]
	\centering
    \includegraphics[width=\linewidth]{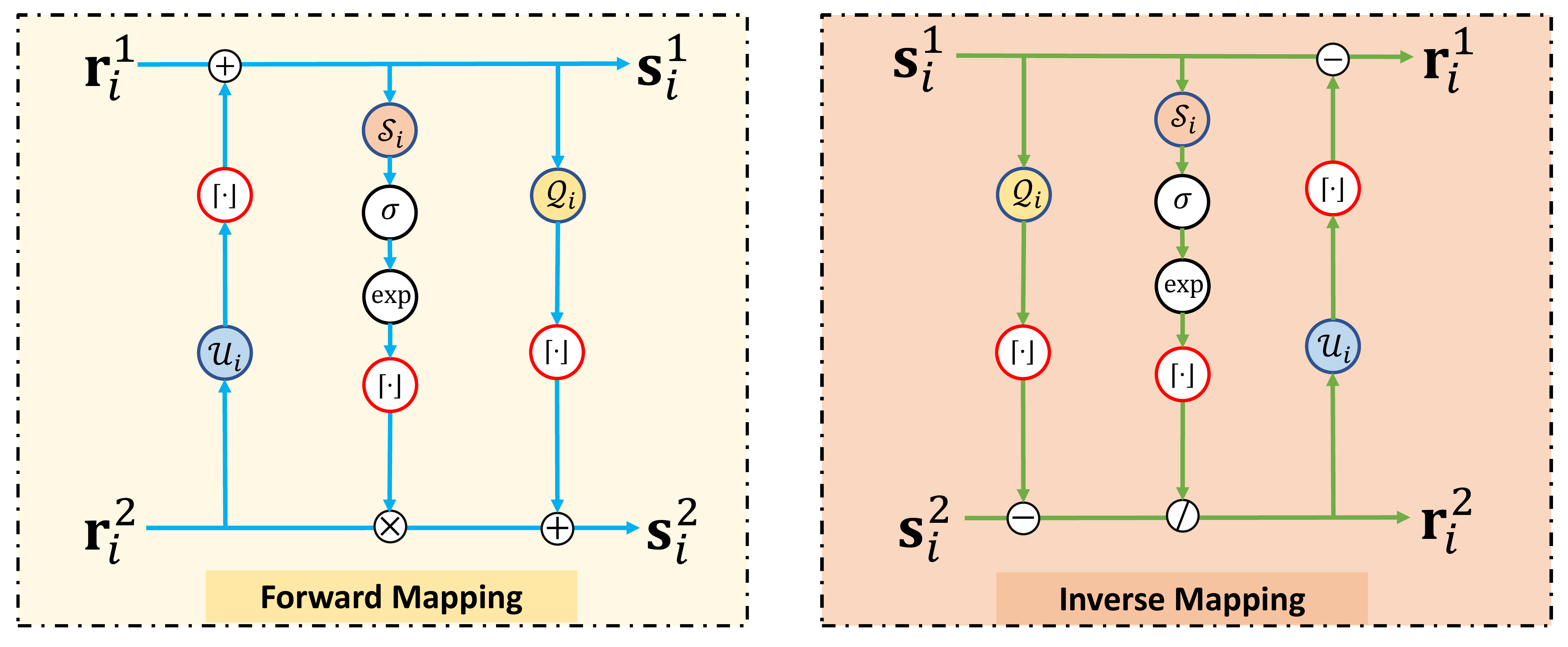}
    \caption{Computation diagram of the forward and inverse mappings in the $i$th integer coupling layer of the integer invertible watermark network.}
    \label{fig:coupling_layer}
\end{figure}

This section introduces the DRRW method, which consists of three key components.
First, we design the integer invertible watermark network (iIWN) with integer coupling layers as shown in Figure~\ref{fig:coupling_layer}. This design ensures lossless forward and inverse mappings for integer-valued data, avoiding irreversible issues caused by quantization errors.
Second, the DRRW uses an encoder-noise layer-decoder training framework as shown in Figure \ref{fig:training}. Compared to existing RRW methods, DRRW can learn robustness against different distortions through a differentiable noise pool.
Third, the DRRW inference process shown in Figure \ref{fig:inference} follows the two-stage embedding, similar to traditional RRW methods.

\vspace{-10pt}
\subsection{Integer Invertible Watermark Network}\label{sec:idinn}
Flow-based networks can be considered as compositions of \(L\) coupling layers, forming an invertible neural network denoted as \(\mathcal{F}_\theta = \mathcal{G}_\theta^1 \circ \mathcal{G}_\theta^2 \circ \cdots \circ \mathcal{G}_\theta^L\), where the inverse function \(\mathcal{F}_\theta^{-1}\) shares the same parameters \(\theta\). The primary distinction between real-valued flows and integer discrete flows lies in their ability to perform invertible mappings: real-valued flows operate between real-valued data distributions (\(\mathbf{x}, \mathbf{y} \in \mathbb{R}\)), while integer discrete flows map between discrete data distributions (\(\mathbf{x}, \mathbf{y} \in \mathbb{Z}\)), where \(\mathbf{y} = \mathcal{F}_{\theta}(\mathbf{x})\).

The coupling layer designed in the flow model has become a fundamental building block for constructing such INNs. The dual-input structure of coupling layers, which can process both the cover image and watermark, enhances their suitability for watermark embedding and extraction. As a result, INN has been widely applied in image steganography and watermarks.
Existing flow-based watermark methods are all based on real-valued flow-based, both the watermark embedding and extraction processes were modeled as an invertible function \(\mathcal{F}_\theta(\cdot)\), with embedding and extraction corresponding to the forward and inverse mappings of the function, respectively. Let \(\mathbf{I}_{\text{org}} \in \{0, 1, \ldots, 255\}^{H \times W \times C}\) represent the cover image, and \(\mathbf{w} \in \{0, 1\}^M\) denote the watermark bits. The watermark embedding process can be written as \([\mathbf{I}_{\text{stego}}, \mathbf{z}] = \mathcal{F}_\theta([\mathbf{I}_{\text{org}}, \mathbf{w}])\), while the extraction process is expressed as \([\mathbf{I}_{\text{rec}}, \hat{\mathbf{w}}] = \mathcal{F}_\theta^{-1}([\mathbf{I}_{\text{stego}}, \hat{\mathbf{z}}])\), where \(\hat{\mathbf{z}}\) sampled from a tractable distribution, \(\mathbf{I}_{\text{rec}}\) is the reconstructed cover image and \(\hat{\mathbf{w}}\) is the extracted watermark bits. 
Although real-valued flow methods have demonstrated performance in image steganography and watermark tasks, they are inherently designed for real-valued data distributions, and can not handle mapping between integer data distributions.

\subsubsection{Integer Coupling Layer}
To address the limitations of real-valued flow-based models, we design the iIWN, which utilizes integer coupling layers to perform invertible mappings between discrete data distributions. As illustrated in Figure~\ref{fig:coupling_layer}, for the \(i\)-th integer coupling layer, the outputs from the preceding layer, denoted as \(\mathbf{r}_i^1\) and \(\mathbf{r}_i^2\), undergo a forward mapping expressed as:
\begin{equation}
\label{eqn:modified_forward_rs}
\begin{aligned}
\mathbf{s}_i^1 &= \mathbf{r}_i^1 + \lfloor \mathcal{U}_i(\mathbf{r}_i^2) \rceil, \\
\mathbf{s}_i^2 &= \mathbf{r}_i^2 \odot \lfloor \exp(\sigma(\mathcal{S}_i(\mathbf{s}_i^1)))\rceil + \lfloor \mathcal{Q}_i(\mathbf{s}_i^1) \rceil,
\end{aligned}
\end{equation}
where \(\odot\) denotes element-wise multiplication, \(\sigma(\cdot)\) is the sigmoid function, and \(\lfloor \cdot \rceil\) represents the rounding operation.
In contrast, the inverse mapping is given by:
\begin{equation}
\label{eqn:modified_inverse_rs}
\begin{aligned}
\mathbf{r}_i^2 &= (\mathbf{s}_i^2 - \lfloor \mathcal{Q}_i(\mathbf{s}_i^1) \rceil) \oslash \lfloor \exp(\sigma(\mathcal{S}_i(\mathbf{s}_i^1)))\rceil, \\
\mathbf{r}_i^1 &= \mathbf{s}_i^1 - \lfloor \mathcal{U}_i(\mathbf{r}_i^2) \rceil,
\end{aligned}
\end{equation}
where \(\oslash\) represents element-wise division, \(\mathbf{s}_i^1\) and \(\mathbf{s}_i^2\) are the outputs of the current layer. Here, $\mathcal{U}_i(\cdot)$ and $\mathcal{S}_i(\cdot)$, $\mathcal{Q}_i(\cdot)$ are the up/down-sampling networks respectively.

It is important to ensure that \( \exp(\sigma(\mathcal{S}_i(\mathbf{s}_i^1))) \notin (-1, 1) \) when designing the coupling layers, particularly for the term \( \mathbf{r}_i^2 \cdot \exp(\sigma(\mathcal{S}_i(\mathbf{s}_i^1))) \). This constraint prevents division-by-zero issues during the inverse mapping in Eqn.~(\ref{eqn:modified_inverse_rs}), thereby avoiding illegal numerical operations.

While the rounding operator (\(\lfloor \cdot \rceil\)) in Eqns.~(\ref{eqn:modified_forward_rs}) and (\ref{eqn:modified_inverse_rs}) ensure a lossless and invertible mapping between the discrete integer data, however, the gradient of the rounding operator is zero almost everywhere, making it impossible to optimize the model parameters via backpropagation. To address this, we adopt the Straight-Through Estimator \cite{bengio2013estimating} to solve this problem. For a variable \(\mathbf{y} = \lfloor \mathbf{x} \rceil\), the gradient with respect to \(\mathbf{x}\) is approximated as \(\nabla_x \lfloor \mathbf{x} \rceil \approx \mathbf{I}\). This can be implemented as:
\begin{equation}
\mathbf{y} = \mathbf{x} + \texttt{stop\_gradient}(\lfloor \mathbf{x} \rceil - \mathbf{x}),
\end{equation}
This approximation resolves the gradient vanishing problem, facilitating effective optimization of the network parameters.

\begin{figure*}[t!]
	\centering
    \includegraphics[width=\linewidth]{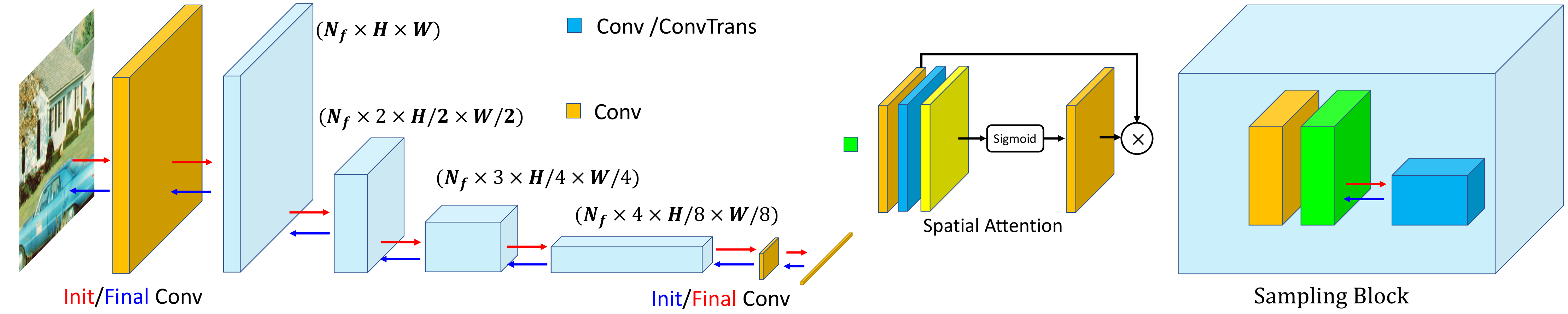}
    \caption{The up/down-sampling network architecture of the DRRW framework. The up-sampling and down-sampling networks exhibit a symmetric structure, where the red arrows represent the down-sampling network and the blue arrows indicate the up-sampling network. Both networks have an initial convolutional layer, multiple sampling blocks, and a final convolutional layer.}
    \label{fig:networks}
    \vspace{-10pt}
\end{figure*}
\subsubsection{The Up/Down-Sampling Network Architecture}
Recognizing that the image and watermark often have significantly different dimensionalities. Therefore, the architecture of the networks \(\mathcal{U}_i(\cdot)\), \(\mathcal{Q}_i(\cdot)\), and \(\mathcal{S}_i(\cdot)\) within the integer coupling layer are different. This distinction arises because dimensional consistency must be preserved when performing numerical operations between elements, ensuring the numerical operations are mathematically valid. In this context, \(\mathcal{U}_i(\cdot)\) usually serves as an up-sampling network to increase the dimensionality of the input features, aligning them with the watermark. Conversely, \(\mathcal{Q}_i(\cdot)\) and \(\mathcal{S}_i(\cdot)\) are designed as down-sampling networks, decreasing the dimensionality for extraction tasks.

In the proposed DRRW framework, the up-sampling and down-sampling networks are structurally symmetric as shown in Figure~\ref{fig:networks}. Both consist of three fundamental components: an initial convolutional layer, multiple sampling blocks, and a final convolutional layer. The initial convolutional layer maps the input image to a fixed number of feature maps, providing essential features for subsequent sampling blocks. Suppose the input image is \(\mathbf{x} \in \mathbb{R}^{C \times H \times W}\), the output features after the initial convolutional layer can be expressed as  
\begin{equation}
    \mathbf{z}_{\text{initial}} = \text{Conv2D}(\mathbf{x}, \mathbf{w}_{\text{initial}}, \mathbf{b}_{\text{initial}}),
\end{equation}
where \(\mathbf{w}_{\text{initial}}, \mathbf{b}_{\text{initial}}\) represents the convolutional weights and the bias terms respectively. The output features \(\mathbf{z}_{\text{initial}} \in \mathbb{R}^{N_f \times H \times W}\) contain \(N_f\) feature channels.

The subsequent layers consist of several down-sampling blocks. Each block is composed of a convolution, a spatial attention module, and a down-sampling convolution. Let \(\mathbf{z}^{\text{l}}_{i-1}\) represent the output of the \((i-1)\)-th down-sampling block. The transformation process within the \(i\)-th sampling block can be expressed as:
\begin{align}
\mathbf{z}_i^{\text{f}} &= \text{Conv2D}(\mathbf{z}^{\text{l}}_{i-1}, \mathbf{w}_i^{\text{f}}, \mathbf{b}_i^{\text{f}}), \\
\mathbf{z}_i^{\text{s}} &= \text{SpatialAttention}(\mathbf{z}_i^{\text{f}}), \\
\mathbf{z}_i^{\text{l}} &= \text{Conv2D}(\mathbf{z}_i^{\text{s}}, \mathbf{w}_i^{\text{l}}, \mathbf{b}_i^{\text{l}}).
\end{align}
It is important to note that the final output feature \(\mathbf{z}_i^{\text{l}}\) has a spatial resolution reduced to half of the original, while the number of feature channels increases by $N_f$. After bypassing through \(N_d\) down-sampling block, the final feature representation is fed into the final convolutional layer to obtain the watermark feature representation:
\begin{equation}
\mathbf{z}_w = \text{Conv2D}(\mathbf{z}^{\text{l}}_{N_d}, \mathbf{w}_l, \mathbf{b}_l),
\end{equation}
where \(\mathbf{z}_w \in \mathbb{R}^{L \times L}\) represents the final watermark feature map.

As shown in Figure~\ref{fig:networks}, the blue arrows represent the up-sampling network. Since the up-sampling network is symmetric to the down-sampling network, the main difference from the down-sampling network is that the down-sampling convolution in the down-sampling network is replaced by a transposed convolution in the up-sampling network, which is used to increase the dimensionality of the features. This enables the up-sampling to extract features for stego images.

\subsection{Training}\label{sec:proposed_msw}
\begin{figure*}[th]
	\centering
    \includegraphics[width=\linewidth]{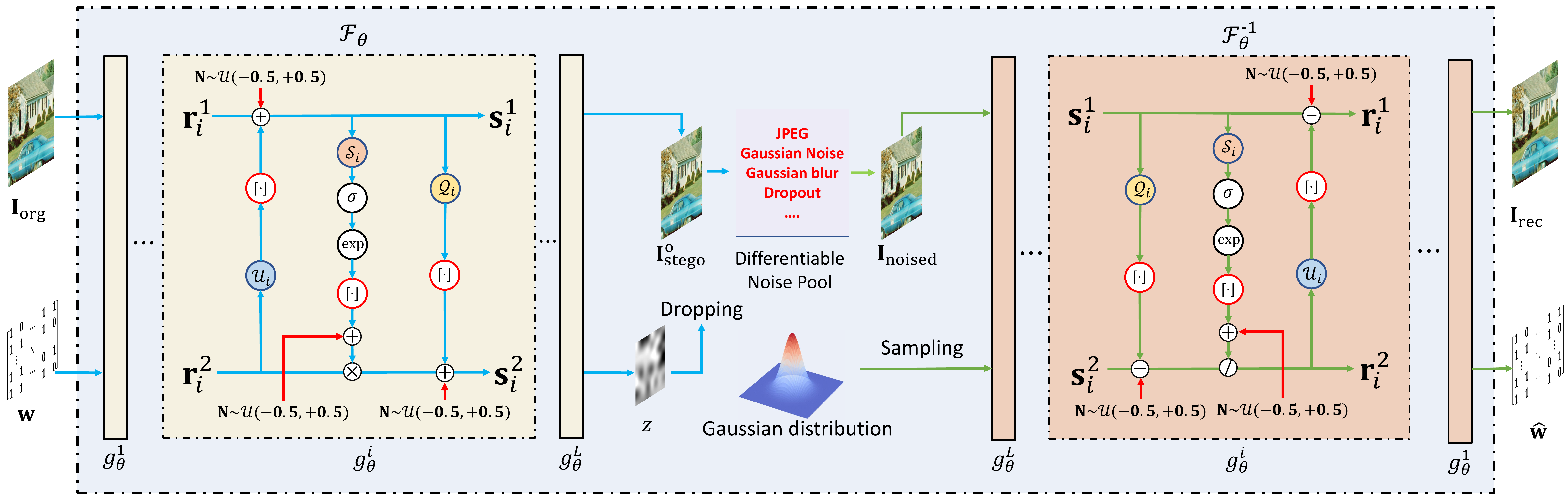}
    \caption{The training framework of DRRW. DRRW adopts the encoder-noise layer-decoder training structure common to other deep learning-based watermarking methods. Unlike traditional robust reversible watermarking approaches, DRRW enhances robustness against various distortions by learning.}
	\label{fig:training}
\end{figure*}
As shown in Figure \ref{fig:training}, in the training process, we adopt the encoder-noise layer-decoder framework: in the forward mapping, the cover image \(\mathbf{I}_{\text{org}}\) and watermark \(\mathbf{w}\) are mapped through \(\mathcal{F}_{\theta}\) to obtain an overflowed stego image \(\mathbf{I}_{\text{stego}}^{\text{o}}\) and a latent variable \(\mathbf{z}\) as \(\left[\mathbf{I}_{\text{stego}}^{\text{o}}, \mathbf{z}\right] = \mathcal{F}_{\theta}(\left[\mathbf{I}_{\text{org}}, \mathbf{w}\right])\). Then, \(\mathbf{I}_{\text{stego}}^{\text{o}}\) undergoes distortions through a noise layer \(\mathcal{N}(\cdot)\), resulting in \(\mathbf{I}_{\text{noised}} = \mathcal{N}(\mathbf{I}_{\text{stego}}^{\text{o}})\). In the reverse mapping, \(\mathbf{I}_{\text{noised}}\) and a latent variable \(\hat{\mathbf{z}}\) sampled from a Gaussian distribution are used with the inverse mapping \(\mathcal{F}_{\theta}^{-1}\) to reconstruct the cover image and extract the watermark bits as \(\left[\mathbf{I}_{\text{rec}}, \hat{\mathbf{w}}\right] = \mathcal{F}_{\theta}^{-1}(\left[\mathbf{I}_{\text{noised}}, \mathbf{z})\right]\).
\subsubsection{Differentiable Noise Pool}
To enhance the robustness of the watermark, the overflowed stego image \(\mathbf{I}_{\text{stego}}^{\text{o}}\) is subsequently imputed into a differentiable noise pool \(\mathcal{N}(\cdot)\). This introduces a variety of distortions to the stego image, producing a noised stego image \(\mathbf{I}_{\text{noised}} = \mathcal{N}(\mathbf{I}_{\text{stego}}^{\text{o}})\). 
The differentiable noise pool includes six common distortions: JPEG compression, Gaussian blur, Gaussian noise, salt\&pepper noise, median filtering, and dropout.

\subsubsection{Loss Functions}  
The loss function for our proposed method consists of forward and reverse losses. The forward loss aims to reduce the number of overflowed pixels while minimizing the difference between the overflowed stego image and the original cover image. The loss related to the stego image is defined as:
\begin{align}
\mathcal{L}_{\text{s}} = \text{MSE}(\mathbf{I}_{\text{stego}}^{\text{o}}, \mathbf{I}_{\text{org}}),~~
\mathcal{L}_{\text{l}} = \text{LPIPS}(\mathbf{I}_{\text{stego}}^{\text{o}}, \mathbf{I}_{\text{org}}). 
\end{align}
where \(\text{LPIPS}(\cdot,\cdot)\) is the learned perceptual image patch similarity \cite{zhang2018lpips}. Furthermore, To suppress pixel overflow in $\mathbf{I}_{\text{stego}}^{\text{o}}$, we propose a penalty loss, which defined as:  
\begin{equation} 
\mathcal{L}_{\text{p}} = \text{MSE}(\mathbf{I}_{\text{stego}}^{\text{+}}, \mathbf{0}) + \text{MSE}(\mathbf{I}_{\text{stego}}^{\text{-}}, \mathbf{0}),
\end{equation}  
where \(\mathbf{I}_{\text{stego}}^{\text{+}} = \texttt{relu}(\mathbf{I}_{\text{stego}}^{\text{o}} - 255)\) and \(\mathbf{I}_{\text{stego}}^{\text{-}} = \texttt{relu}(\mathbf{0} - \mathbf{I}_{\text{stego}}^{\text{o}})\). The \(\texttt{relu}(\cdot)\) function suppresses pixel values that exceed the valid range \([0, 255]\), and by penalizing pixels outside this range, it not only solves overflow and underflow issues but also improves the visual quality of stego image.
In addition, we introduce a regularization loss term on the latent variable \(\mathbf{z}\), defined as follows:
\begin{equation} 
\mathcal{L}_{\text{z}} = \text{MSE}(\mathbf{z}, \mathbf{0}).
\end{equation}
This regularization loss penalizes excessively large network parameters that lead to numerical overflow during inference, and it also reduces the length of the encoded bitstream. Experimental analysis is provided in Section~\ref{sec:ablation}.

In the reverse mapping, the objective is to improve the accuracy of watermark extraction. The loss for watermark extraction is formulated as:  
\begin{equation} 
\mathcal{L}_{\text{w}} = \text{MSE}(\hat{\mathbf{w}}, \mathbf{w}), 
\end{equation}  
where \(\hat{\mathbf{w}},\mathbf{w}\) are the extracted watermark and the original watermark, respectively.

Combining all the loss functions, the overall loss function is formulated as:  
\begin{equation} 
\mathcal{L} = \lambda_s \cdot \mathcal{L}_{\text{s}} + \lambda_l \cdot \mathcal{L}_{\text{l}} + \lambda_z \cdot \mathcal{L}_{\text{z}} + \lambda_p \cdot \mathcal{L}_{\text{p}} + \lambda_w \cdot \mathcal{L}_{\text{w}}.
\end{equation}  
where \(\lambda_s\), \(\lambda_l\), \(\lambda_z\), \(\lambda_p\), and \(\lambda_w\) represent weighting coefficients that balance the contribution of each loss term.

\subsubsection{Training Strategy}\label{sec:train_strategy}
During training, different loss terms often require careful weighting to stabilize optimization. However, tuning these weights is typically labor-intensive and computationally costly. To address this, we propose an adaptive loss weight adjustment strategy.

In watermarking tasks, the primary goal is to embed the watermark and accurately extract it, while preserving image quality. Importantly, watermark extraction accuracy is often prioritized over image imperceptibility during training. Therefore, \textit{can we adjust the watermark loss weight to gradually balance the watermark extraction accuracy and imperceptibility by fixing the weights of other losses?} We consider this idea is feasible. 
Specifically, we initialize the watermark loss weight with a large value and adaptively adjust it based on extraction accuracy over recent epochs. 

Let $\mathbf{A} = [a_{t-n}, \dots, a_{t}]$ denote the extraction accuracies from epoch $t - n$ to $t$. The average accuracy is computed as:
\[
\text{acc} = \frac{1}{n}\sum_{i=t-n}^{t} \mathbf{A}(i),
\]
where $n$ is the average window size. When $\text{acc}$ exceeds a target threshold or the extraction error falls below a preset tolerance $\delta$, it indicates stable watermark embedding under the current weight \(\lambda_w^{t}\). We then reduce the weight to enhance image quality:
\[
\lambda_{\text{w}}^{t+1} =\lambda_{\text{w}}^t \times v, ~~~\text{if} \ \text{acc} > 1 - \delta,
\]
where $\delta \in (0, 1)$ is the tolerable error, and \(v \in (0, 1)\) is a discount factor.
This adaptive adjustment stabilizes training and facilitates a progressive balance between watermark robustness and image quality.
Furthermore, we adopt the stochastic round for training. As shown in Figure \ref{fig:training}, The quantized outputs are added with a uniformly distributed noise as \(\mathbf{y} = \mathbf{x} + \texttt{stop\_gradient}(\lfloor \mathbf{x} \rceil - \mathbf{x}) + \mathbf{u}\), where \(\mathbf{u}\sim U(\mathbf{-0.5}, \mathbf{+0.5})\).

\subsection{Inference}
\label{sec:inference}
\begin{figure*}[t]
    \centering
    \includegraphics[width=\linewidth]{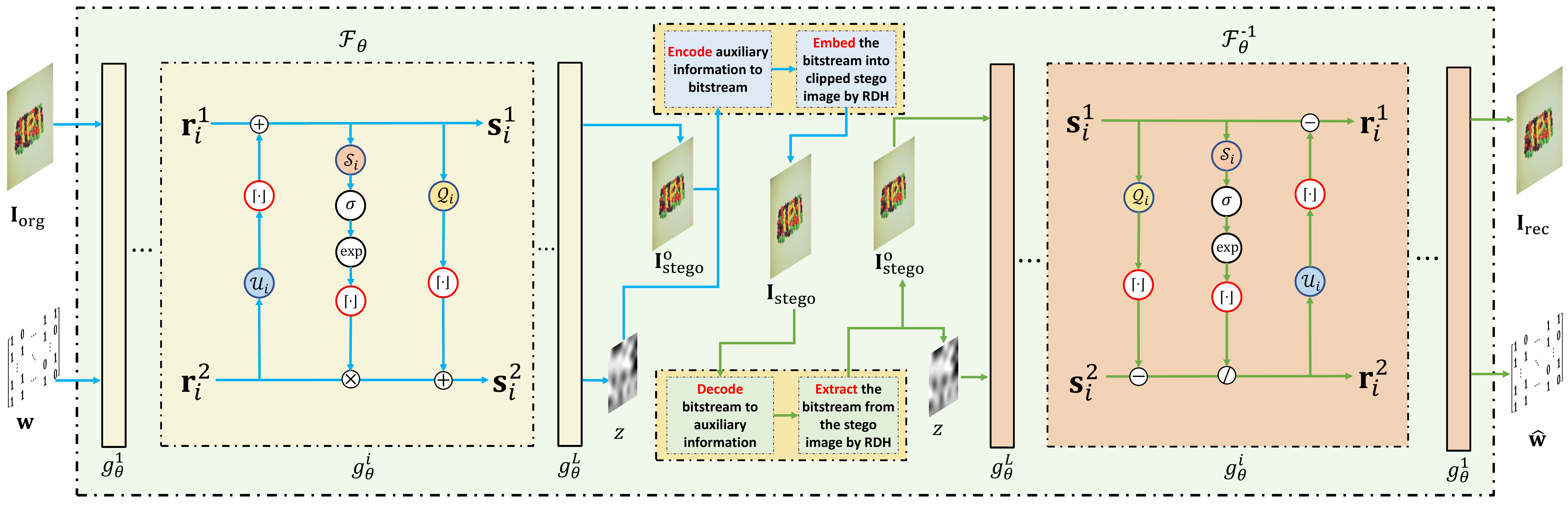}
    \caption{Overview of the inference process in DRRW. DRRW employs a two-stage embedding strategy. In the first stage, the cover image-watermark pair is forward-mapped into an overflowed stego image and latent variables through an integer invertible watermark network. In the second stage, auxiliary bitstreams are embedded into the clipped stego image using reversible data hiding.}
    \label{fig:inference}
\end{figure*}
The inference phase proceeds as shown in Figure~\ref{fig:inference}. The overflowed stego image generated through forward mapping via the iIWN is similar to the process in the training stage. Simultaneously, the overflow map and latent variables are losslessly compressed into a bitstream using arithmetic coding \cite{witten1987arithmetic}. This bitstream is then embedded into the stego image using an RDH \cite{sachnev2009reversible}.

\subsubsection{Embedding Process}
During the embedding process, two primary challenges must be addressed: (1) storing overflowed pixels in the stego image, and (2) storing the latent variable. For the overflowed pixels, we extract the overflowed portion as an overflow map \(\mathbf{O}\), defined as:
\begin{equation}
    \mathbf{O}(i,j,c) = \begin{cases} 
        \mathbf{I}_{\mathrm{stego}}^{\text{o}}(i,j,c) - 255, & \text{if } \mathbf{I}_{\mathrm{stego}}^{\text{o}}(i,j,c) > 255, \\
        0 - \mathbf{I}_{\mathrm{stego}}^{\text{o}}(i,j,c), & \text{if } \mathbf{I}_{\mathrm{stego}}^{\text{o}}(i,j,c) < 0, \\
        0, & \text{otherwise}.
    \end{cases}
\end{equation}
Here, pixel values within the valid range are recorded as zeros in \(\mathbf{O}\), while overflowed pixel values are stored as their overflow amounts in \(\mathbf{O}\). Both \(\mathbf{O}\) and the latent variable \(\mathbf{z}\) are encoded into a bitstream using arithmetic coding. This bitstream is then embedded into \(\mathbf{I}_{\mathrm{stego}}^{\mathrm{clip}} = \mathrm{clip}(\mathbf{I}_{\mathrm{stego}}^{\text{o}}, 0, 255)\) via RDH, resulting in the final stego image \(\mathbf{I}_{\mathrm{stego}}\).

\subsubsection{Watermark Extraction and Lossless Cover Recovery}
Watermark extraction and lossless cover recovery are performed over lossy and lossless channels, respectively.

\paragraph{Lossy Channels} For the lossy channels, the primary goal is robust watermark extraction. Given a noised stego image $\mathbf{I}_{\mathrm{noised}}$, the watermark is extracted via inverse mapping, \emph{i.e.,} \(\left[\mathbf{I}_{\text{rec}}, \hat{\mathbf{w}}\right] = \mathcal{F}_{\theta}^{-1}(\left[\mathbf{I}_{\text{noised}}, \mathbf{z})\right]\).

\paragraph{Lossless Channels}
For lossless channels, our primary goal is to recover the original cover image and watermark bit. First, we extract the auxiliary bitstream from the stego image \(\mathbf{I}_{\mathrm{stego}}\) using the RDH method. Then, we decode the auxiliary bitstream with arithmetic coding to retrieve the overflow map \(\mathbf{O}\) and latent variable \(\mathbf{z}\). Subsequently, the overflowed stego image \(\mathbf{I}_{\mathrm{stego}}^{\text{o}}\) is reconstructed as:
\begin{equation}
    \mathbf{I}_{\mathrm{stego}}^{\text{o}} = \mathbf{I}_{\mathrm{stego}}^{\mathrm{clip}} + \mathbf{I}_{\mathrm{mask}}^{255} \odot \mathbf{O} - \mathbf{I}_{\mathrm{mask}}^{0} \odot \mathbf{O},
\end{equation}
where the masks are defined as:
\begin{align}
    \mathbf{I}_{\mathrm{mask}}^{255}(i,j,c) &= \begin{cases} 1, & \text{if } \mathbf{O}(i,j,c) > 0, \\ 0, & \text{otherwise}, \end{cases} \\
    \mathbf{I}_{\mathrm{mask}}^{0}(i,j,c) &= \begin{cases} 1, & \text{if } \mathbf{O}(i,j,c) < 0, \\ 0, & \text{otherwise}. \end{cases}
\end{align}
It should be noted that the clipped image \(\mathbf{I}_{\mathrm{stego}}^{\mathrm{clip}}\) is recovered via RDH. Finally, the original cover image and watermark are recovered through inverse mapping, ensuring perfect recovery of both the original cover image \(\mathbf{I}_{\mathrm{org}}\) and the watermark \(\mathbf{w}\).

In summary, DRRW enables robust watermark extraction in lossy channels and perfect recovery of the cover image and watermark in lossless channels.

\section{Experimental Results}\label{sec:exp}
The experimental analysis consists of four parts. Section \ref{sec:exp_setup} details the experimental settings, including descriptions of the training and testing datasets, implementation details, and hyperparameter configurations.
Section \ref{sec:compare_rrw} conducts a comparative evaluation between the proposed DRRW framework and SOTA two-stage RRW methods. The evaluation assesses watermark imperceptibility, robustness under various distortions, bitstream length required for recovery, and computational complexity.
Section \ref{sec:compare_irw} provides a comparative analysis with irreversible robust watermarking methods, demonstrating that DRRW not only offers competitive robustness but also enables cover image recovery in lossless channels. 
Section \ref{sec:ablation} presents ablation studies to investigate the contributions of components within the DRRW framework.  
Overall, experimental results demonstrate that the proposed DRRW framework exhibits significant and irreplaceable advantages over existing methods, representing a critical step toward the practical deployment of RRW.

\subsection{Experimental Setups} \label{sec:exp_setup}
\subsubsection{Training and Test Datasets}
In all experiments, the training set for all the learning-based methods is the DIVK dataset~\cite{Agustsson_2017_CVPR_Workshops}. The testing set primarily consists of two datasets: the classical USC-SIPI~\cite{uscsipidataset} and Kodak24~\cite{kodakdataset}.

\subsubsection{Implementation Details}
All comparison methods are trained on an NVIDIA 4090 Ti GPU using the same training set as DRRW. The DRRW framework is implemented in PyTorch and optimized with AdamW (batch size \(8\), \(\gamma = 10^{-4}\), \(\beta_1 = 0.5\), \(\beta_2 = 0.999\)). It uses \(L = 5\) integer coupling layers, with \(\delta = 10^{-2}\), \(v = 0.75\), \(\lambda_p = 10^6\), \(\lambda_w = 10^4\), \(\lambda_s = 1\), \(\lambda_l = 5\), \(\lambda_z = 10^{-3}\). In the second stage, arithmetic coding~\cite{witten1987arithmetic} and PEE~\cite{sachnev2009reversible} enable lossless compression and reversible embedding of auxiliary information, respectively. Details for other methods are provided in their respective comparative sections.

\begin{figure*}[t!]
    \centering
    \includegraphics[width=\linewidth]{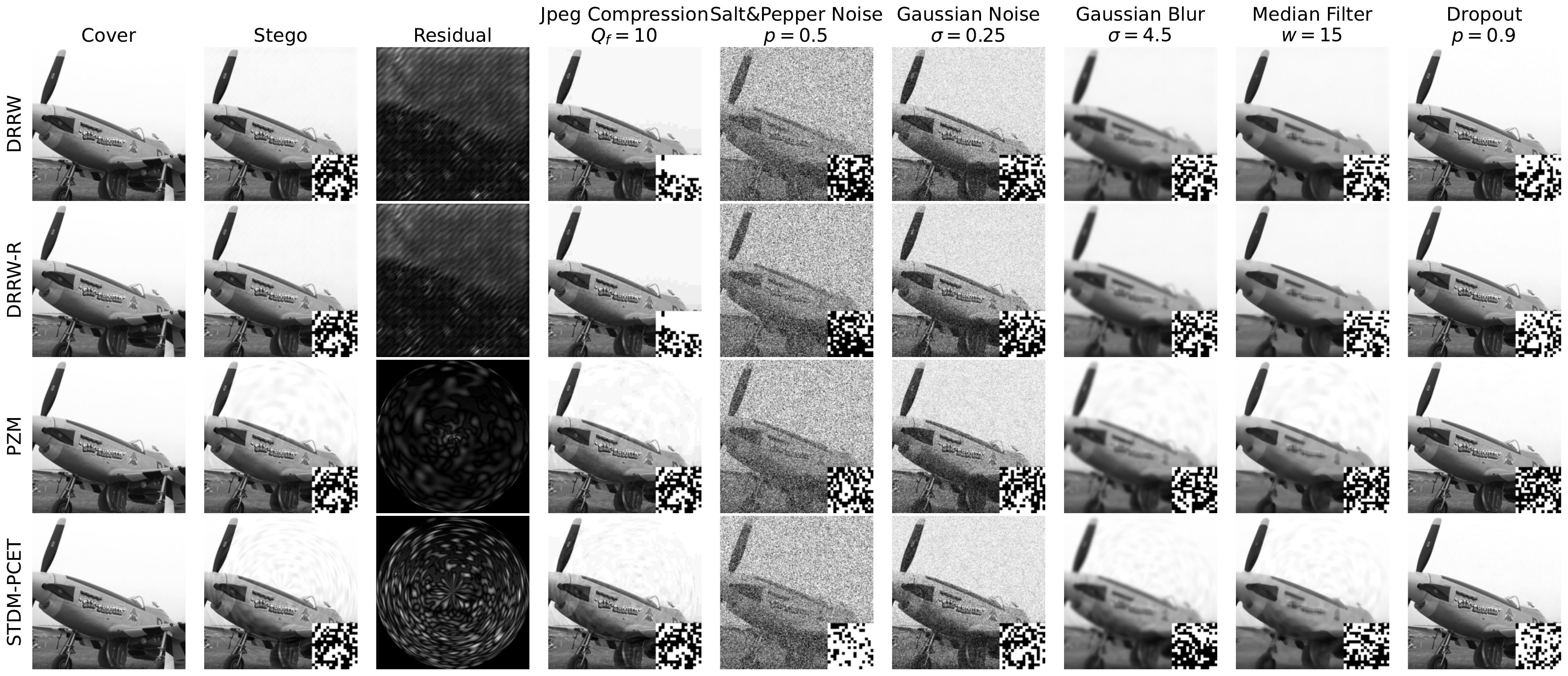}
    \caption{Visual quality comparison between DRRW and STDM-PCET~\cite{tang2024robust}, PZM~\cite{tang2022highly} under various distortions. The residual images demonstrate that DRRW achieves stronger coupling between the stego and cover images, leading to better imperceptibility compared to traditional RRW methods.}
    \label{fig:vq_noise_compare}
\end{figure*}

\begin{figure}[t]
    \centering
    \begin{minipage}[]{.48\linewidth}
        \centering
        \includegraphics[width=\linewidth]{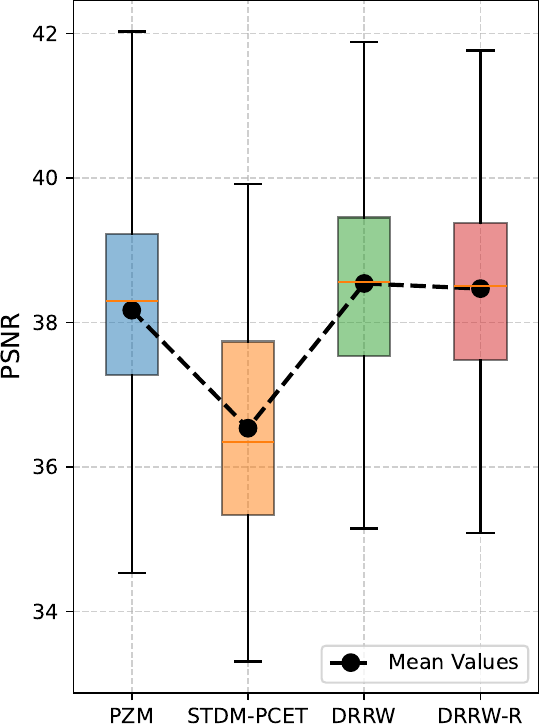}
    \end{minipage}
    \begin{minipage}[]{.5\linewidth}
        \centering
        \includegraphics[width=\linewidth]{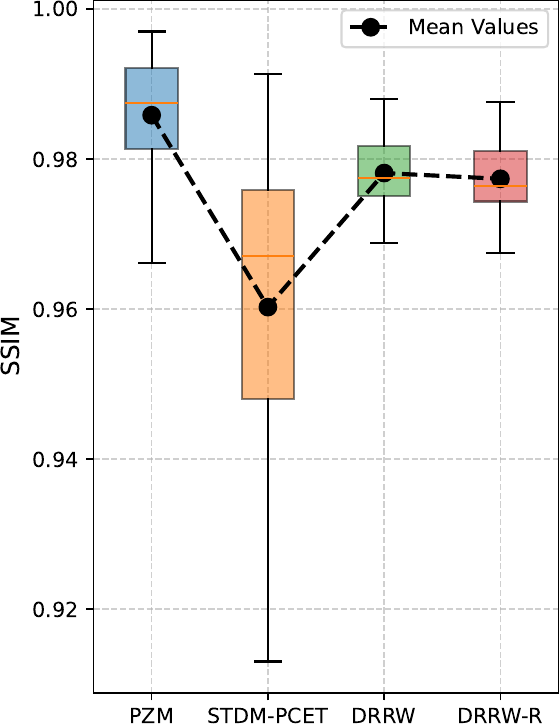}
    \end{minipage}
    \caption{Quantitative visual quality comparison between the proposed DRRW and the STDM-PCET~\cite{tang2024robust}, PZM~\cite{tang2022highly}.}
    \label{fig:rrw_psnr_ssim}
\end{figure}

\subsection{DRRW vs. Traditional Robust Reversible Watermarking}\label{sec:compare_rrw}
This section evaluates the proposed two-stage DRRW method against the two-stage RRW methods STDM-PCET~\cite{tang2024robust} and PZM~\cite{tang2022highly}, examining four critical aspects: visual quality, watermark robustness, time complexity, and auxiliary bitstream length. To ensure fairness, experimental setups are meticulously aligned. DRRW is trained on the DIV2K dataset~\cite{Agustsson_2017_CVPR_Workshops}, with images resized to \(512 \times 512\), converted to grayscale, and embedded with a 256-bit bitstream. STDM-PCET~\cite{tang2024robust} and PZM~\cite{tang2022highly} adopt identical configurations. DRRW and DRRW-R denote stego images from the first and second stages, respectively. Experimental results follow below.
 \begin{figure}[t]
    \centering
    \includegraphics[width=\linewidth]{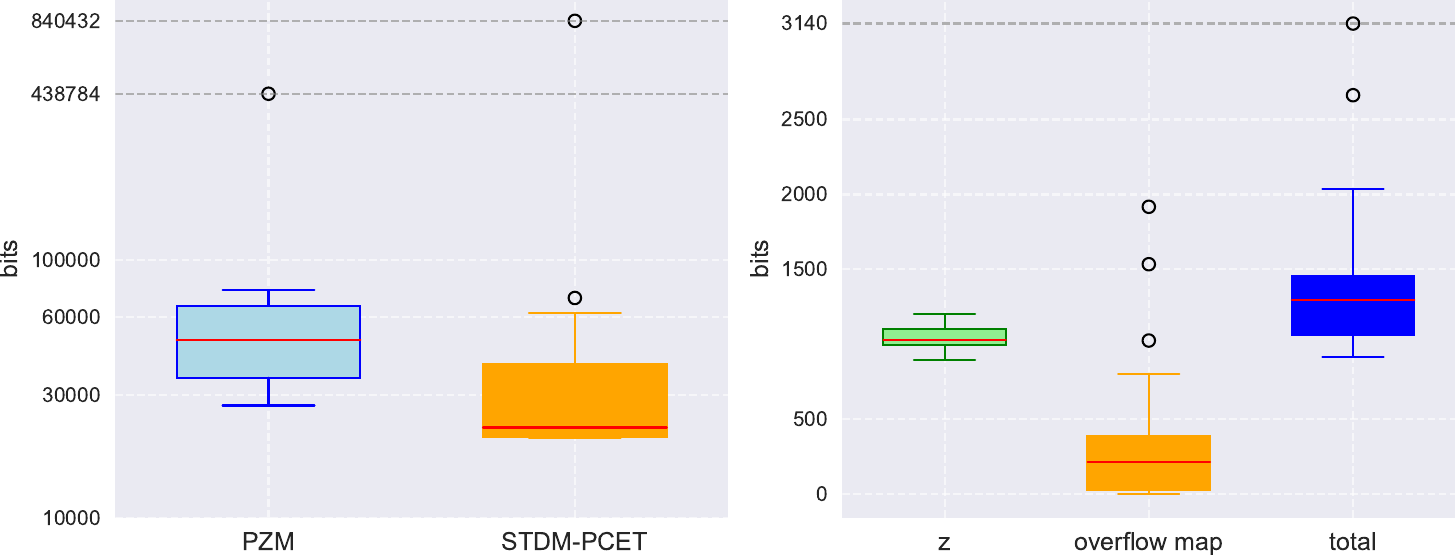}
    \caption{Comparison of auxiliary bitstream lengths: STDM-PCET~\cite{tang2024robust} and PZM~\cite{tang2022highly} (left) versus proposed DRRW (right).}
    \label{fig:auxlens_times}
    \vspace{-10pt}
\end{figure}

\subsubsection{Visual Quality Comparison}
Figure~\ref{fig:vq_noise_compare} compares visual examples of the proposed DRRW with PZM~\cite{tang2022highly} and STDM-PCET~\cite{tang2024robust}. Residual images show that DRRW strengthens watermark-cover image coupling, unlike STDM-PCET and PZM, which rely on template-based embedding, weaken coupling, and produce more artifacts. We assess visual quality using PSNR and SSIM, as summarized in Figure~\ref{fig:rrw_psnr_ssim}. DRRW and DRRW-R yield average PSNR/SSIM of \(38.54/0.978\) and \(38.47/0.977\), outperforming PZM (\(38.17/0.986\)) and STDM-PCET (\(36.54/0.960\)). DRRW boosts PSNR by \(2.00\) dB and SSIM by \(0.018\) compared to STDM-PCET, and increases PSNR by \(0.37\) dB over PZM, demonstrating enhanced image quality and imperceptibility.

\begin{figure*}[t]
    \centering
    \begin{minipage}[]{.32\linewidth}
        \centering
        \includegraphics[width=\linewidth]{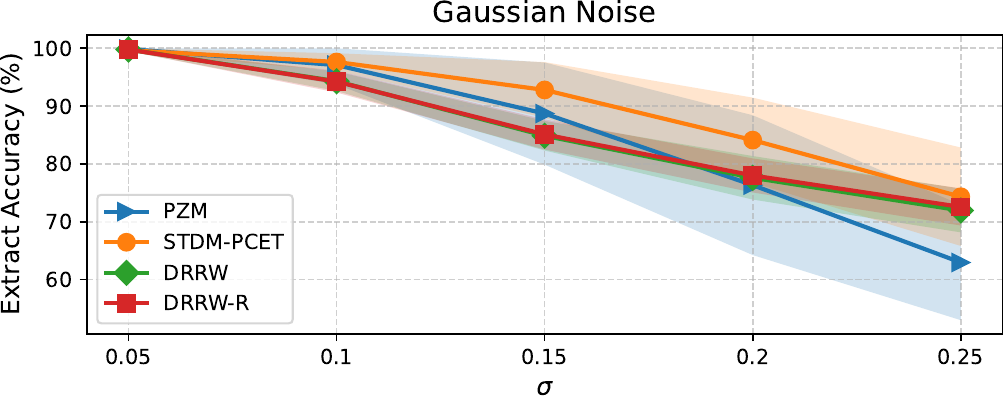}
    \end{minipage}
    \hspace{2pt}
    \begin{minipage}[]{.32\linewidth}
        \centering
        \includegraphics[width=\linewidth]{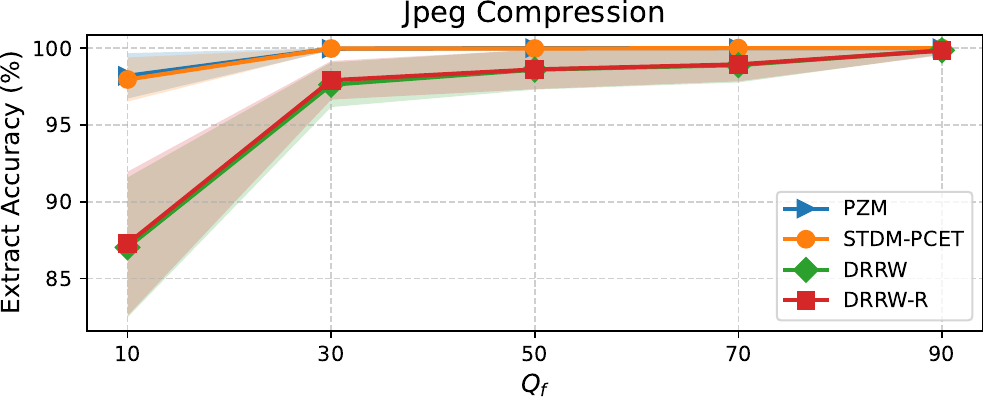}
    \end{minipage}
    \hspace{2pt}
    \begin{minipage}[]{.32\linewidth}
        \centering
        \includegraphics[width=\linewidth]{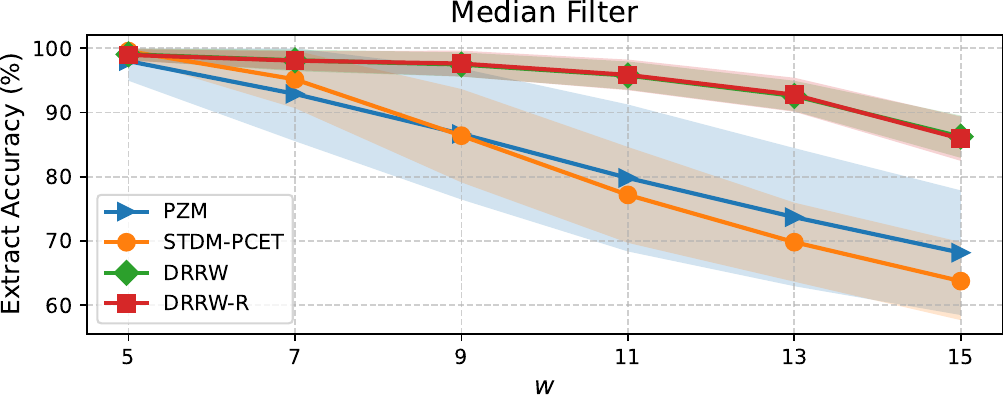}
    \end{minipage}
    \\
    \begin{minipage}[]{.32\linewidth}
        \centering
        \includegraphics[width=\linewidth]{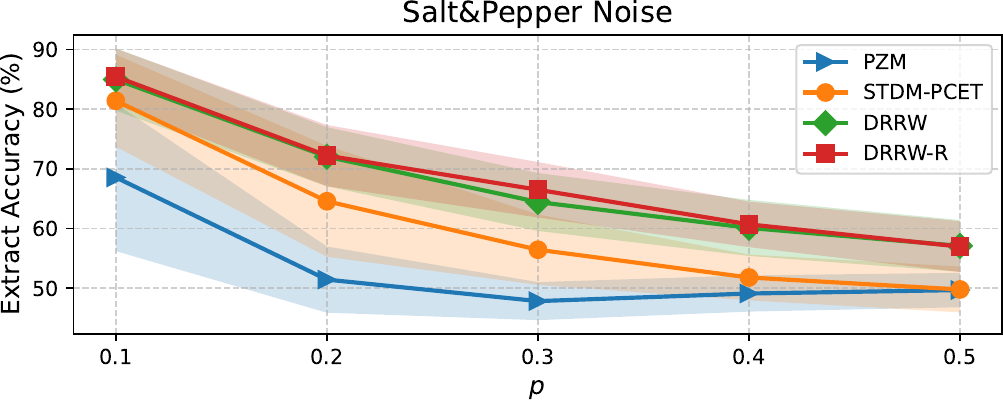}
    \end{minipage}
    \hspace{2pt}
    \begin{minipage}[]{.32\linewidth}
        \centering
        \includegraphics[width=\linewidth]{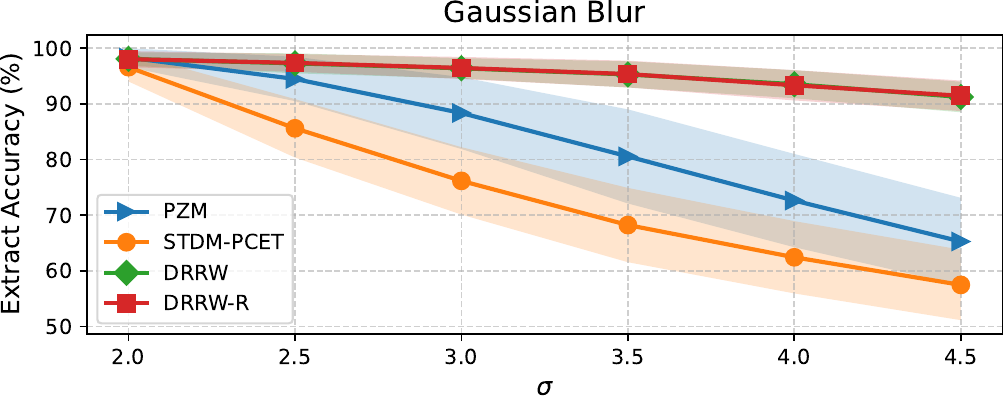}
    \end{minipage}
    \hspace{2pt}
    \begin{minipage}[]{.32\linewidth}
        \centering
        \includegraphics[width=\linewidth]{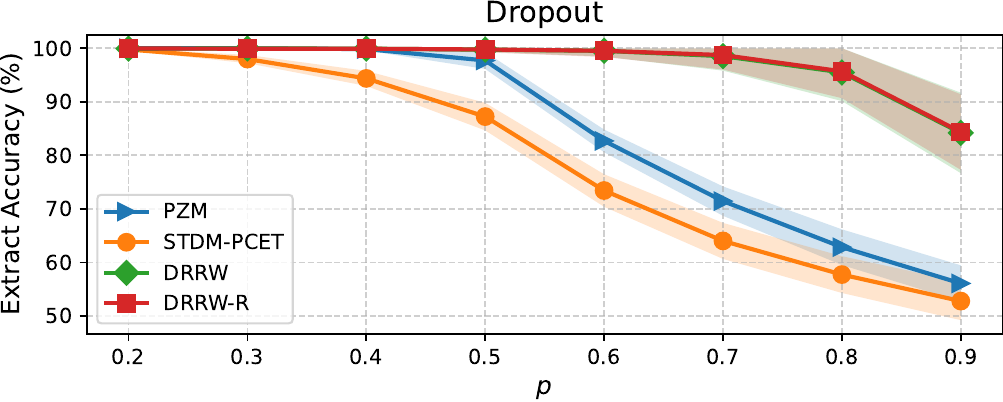}
    \end{minipage}
    \caption{Extraction accuracy (\%) comparison between STDM-PCET~\cite{tang2024robust}, PZM\cite{tang2022highly} and the proposed DRRW under various distortions, including Dropout, Gaussian blur, median filtering, salt-and-pepper noise, Gaussian noise, and JPEG compression. DRRW exhibits superior robustness against spatial-domain distortions while maintaining competitive performance under noise-based attacks.}
    \label{fig:robustness_rrw}
\end{figure*}

\subsubsection{Robustness Comparison}
Six common types of distortions test robustness, with extraction accuracy for each distortion shown in Figure~\ref{fig:robustness_rrw}. Experimental results reveal that the proposed DRRW method outperforms PZM~\cite{tang2022highly} and STDM-PCET~\cite{tang2024robust} in most distortions. For example, under Dropout distortion with \(p=0.9\), DRRW achieves an extraction accuracy of \(84.19\%\), far surpassing PZM’s \(56.07\%\) and STDM-PCET’s \(52.77\%\). With Gaussian blur (\(\sigma=4.5\)), DRRW reaches \(91.24\%\), significantly exceeding PZM’s \(65.26\%\) and STDM-PCET’s \(57.46\%\). Median filtering (\(w=15\)) yields \(86.24\%\) for DRRW, compared to PZM’s \(68.18\%\) and STDM-PCET’s \(63.76\%\). For salt-and-pepper noise (\(p=0.5\)), DRRW attains \(57.11\%\), outpacing PZM’s \(49.71\%\) and STDM-PCET’s \(49.81\%\). However, under high JPEG compression ratio (\(Q_f=10\)), PZM and STDM-PCET outperform DRRW with \(98.20\%\) and \(97.96\%\), respectively, yet DRRW still maintains a robust \(87.03\%\); when \(Q_f > 30\), DRRW’s accuracy nears \(100\%\). For Gaussian noise, PZM, STDM-PCET, and DRRW show similar robustness; at \(\sigma=0.25\), STDM-PCET achieves \(74.33\%\), PZM \(62.97\%\), and DRRW \(71.95\%\). Overall, DRRW exhibits superior robustness against most distortions compared to SOTA RRW methods.
\begin{figure}[t]
    \centering
    \includegraphics[width=\linewidth]{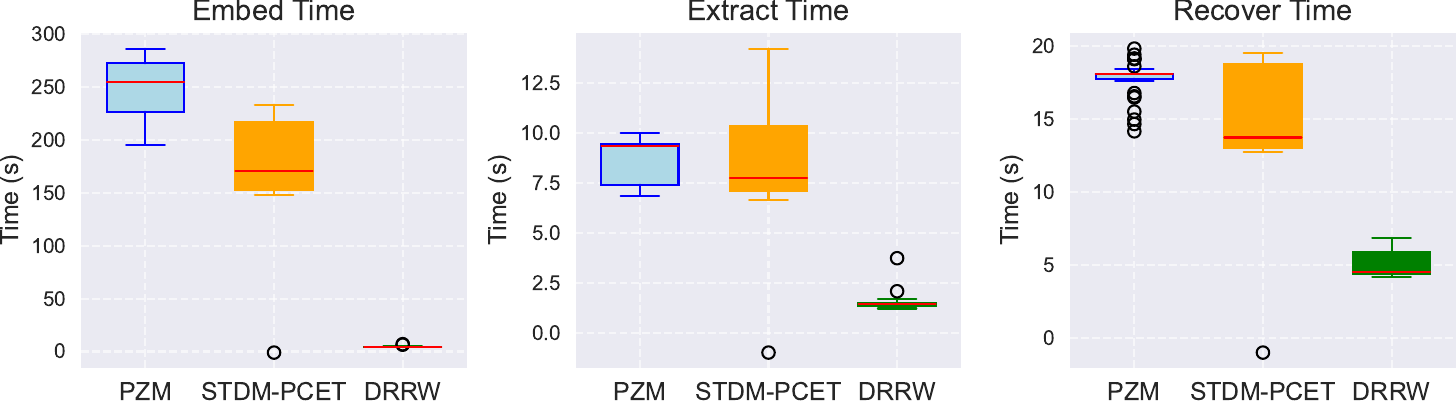}
    \caption{Time complexity and auxiliary bitstream length comparison between DRRW and STDM-PCET~\cite{tang2024robust} and PZM\cite{tang2022highly}.}
    \label{fig:auxlens_times}
\end{figure}
\subsubsection{Time Complexity Comparison}
Figure~\ref{fig:auxlens_times} illustrates the time complexity associated with the embedding, extraction, and recovery processes for STDM-PCET~\cite{tang2024robust}, PZM~\cite{tang2022highly}, and the proposed DRRW. As depicted, PZM requires \(195.17\) to \(286.40\) seconds per sample for embedding, averaging \(248.65\) seconds, while STDM-PCET takes \(147.75\) to \(232.90\) seconds, averaging \(184.26\) seconds. In contrast, DRRW completes embedding within \(4.04\) to \(6.94\) seconds per cover image, averaging \(4.51\) seconds, achieving a \textbf{55.14$\times$} speedup over PZM and \textbf{40.86$\times$} over STDM-PCET. This demonstrates DRRW’s superior embedding efficiency for large-scale applications. For extraction, PZM processes each cover image in \(6.86\) to \(10.01\) seconds, averaging \(8.54\) seconds, and STDM-PCET ranges from \(6.65\) to \(14.24\) seconds, averaging \(8.93\) seconds. DRRW achieves extraction within \(1.21\) to \(3.73\) seconds, averaging \(1.50\) seconds, yielding a \textbf{5.69$\times$} speedup over PZM and \textbf{5.95$\times$} over STDM-PCET. Regarding recovery, PZM averages \(17.71\) seconds with a range of \(14.13\) to \(19.80\) seconds, STDM-PCET ranges from \(12.70\) to \(20.12\) seconds, averaging \(15.89\) seconds, while DRRW completes recovery within \(4.16\) to \(6.83\) seconds, averaging \(4.96\) seconds—a \textbf{3.57$\times$} reduction over PZM and \textbf{3.20$\times$} over STDM-PCET. Overall, DRRW significantly reduces computational costs across all stages, ensuring efficiency in applications on large-scale datasets.

\subsubsection{Auxiliary Bitstream Length Comparison}
The length of the auxiliary bitstream is a crucial factor that directly impacts the reversibility of robust reversible watermarking methods. This is because an excessively long bitstream may prevent reversible embedding in the second stage. Figure~\ref{fig:auxlens_times} illustrates the auxiliary bitstream length required for recovery in STDM-PCET~\cite{tang2024robust}, PZM~\cite{tang2022highly}, and the proposed DRRW. The auxiliary bitstream length in PZM fluctuates widely, spanning \(27{,}248\) to \(438{,}784\) bits, averaging \(61{,}896.94\) bits. STDM-PCET exhibits greater variability, ranging from \(20{,}480\) to \(840{,}432\) bits, with an average of \(56{,}968.52\) bits. In contrast, DRRW significantly reduces the auxiliary bitstream length, ranging from \(919\) to \(3{,}140\) bits, with an average of \(1{,}411.32\) bits, achieving \textbf{43.86$\times$} and \textbf{40.37$\times$} reductions compared to PZM and STDM-PCET, respectively. More specifically, encoding the latent variable \(\mathbf{z}\) requires an average of \(1{,}046.03\) bits, while the overflow map accounts for \(344.29\) bits on average. Notably, for images with limited grayscale variations, such as those with extensive regions near pixel values of \(255\) or \(0\), traditional RRW methods often produce overly long auxiliary bitstreams, rendering reversibility impractical. For example, in the sample shown in Figure~\ref{fig:vq_noise_compare}, the auxiliary bitstream length reaches \(438{,}784\) bits for PZM and \(840{,}432\) bits for STDM-PCET, whereas DRRW requires only \(3{,}140\) bits. Moreover, DRRW successfully achieves reversible embedding for all cover images, without failures. These results underscore the significant advantage of DRRW in reducing auxiliary bitstream length, thus improving practicality.

In summary, the above comparative analysis demonstrates the overall superiority of the proposed DRRW framework over STDM-PCET~\cite{tang2024robust} and PZM~\cite{tang2022highly}. DRRW consistently achieves higher visual quality with fewer embedding artifacts and exhibits stronger robustness against various distortions. Furthermore, DRRW significantly reduces the computational complexity of embedding, extraction, and recovery, achieving up to \textbf{55.14$\times$} speedup during embedding. Additionally, DRRW drastically minimizes the auxiliary information bitstream length, yielding \textbf{43.86$\times$} and \textbf{40.37$\times$} reductions compared to PZM and STDM-PCET, respectively. These results demonstrate that DRRW offers a more practical and reliable solution for robust reversible watermarking by enhancing imperceptibility, robustness, efficiency, and practicality.

\begin{figure*}[t!]
    \centering
    \includegraphics[width=\linewidth]{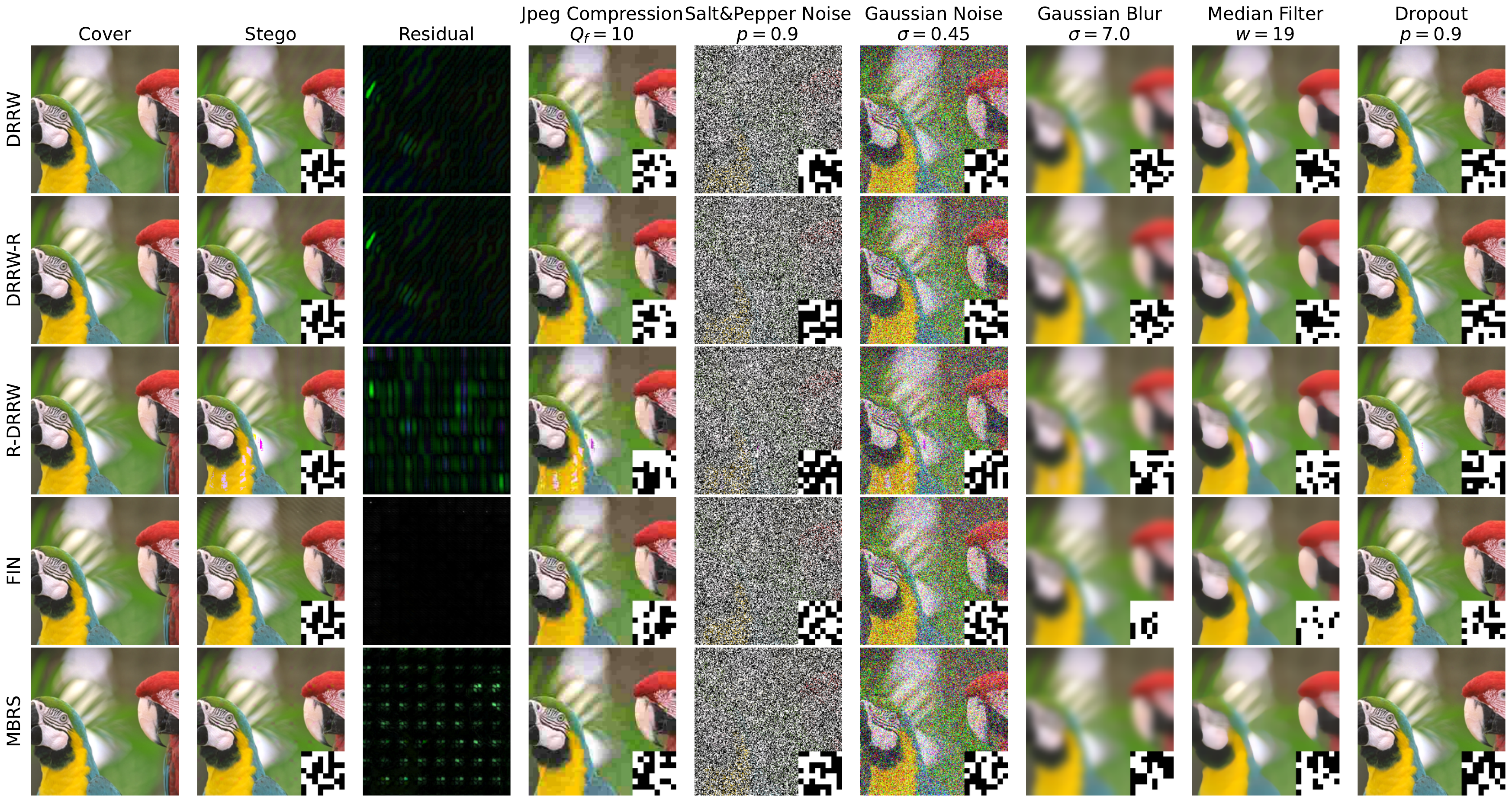}
    \caption{Visual quality comparison between DRRW and the MBRS~\cite{tang2024robust}, FIN~\cite{fang2023flow}, R-DRRW under various distortions. The residual images demonstrate that DRRW achieves stronger coupling between the stego and cover images, leading to better imperceptibility compared to other methods.}
    \label{fig:realflow_vq_compare}
\end{figure*}

\begin{figure}[t]
    \centering
    \begin{minipage}[]{.48\linewidth}
        \centering
        \includegraphics[width=\linewidth]{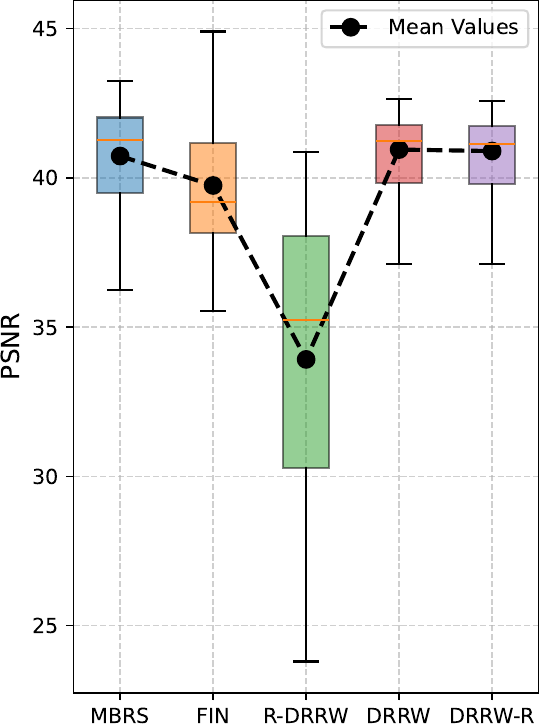}
    \end{minipage}
    \begin{minipage}[]{.48\linewidth}
        \centering
        \includegraphics[width=\linewidth]{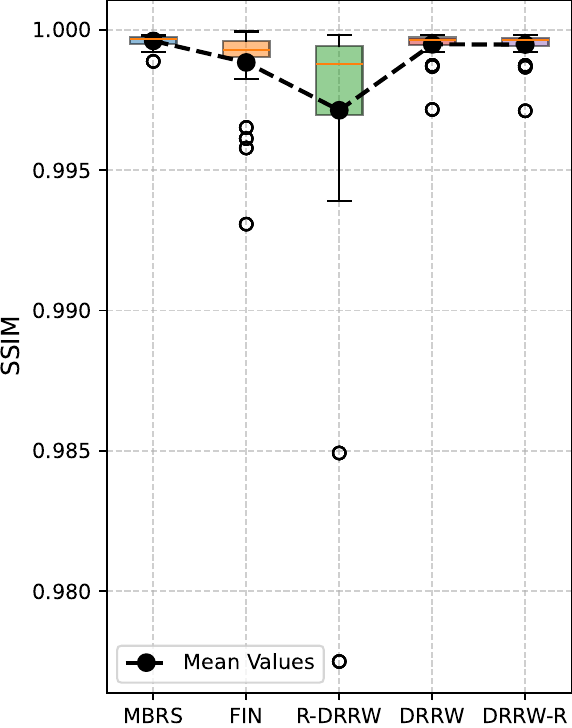}
    \end{minipage}
    \caption{Quantitative visual quality comparison between the proposed DRRW and the MBRS~\cite{tang2024robust}, FIN~\cite{fang2023flow}, R-DRRW in terms of PSNR and SSIM.}
    \label{fig:realflow_psnr_ssim}
\end{figure}

\subsection{DRRW vs. Irreversible Robust Image Watermarking}\label{sec:compare_irw}
This section evaluates DRRW against irreversible robust watermarking methods. The evaluation focuses on visual quality and robustness. Three methods for comparison include real-valued flow-based FIN~\cite{fang2023flow}, MBRS~\cite{jia2021mbrs}, and real-valued DRRW (R-DRRW). FIN and R-DRRW utilize the differentiable noise layer of DRRW for training. R-DRRW deviates from DRRW by removing the quantization operator in the coupling layer and adopting a real-valued flow while keeping other settings identical. For fairness, all methods train on the DIV2K dataset~\cite{Agustsson_2017_CVPR_Workshops}, with images resized to \(256\times 256\) and watermark bit length set to \(64\).

\subsubsection{Visual Quality Comparison}
Figure~\ref{fig:realflow_vq_compare} visualizes stego images from different methods. We observe that the MBRS method~\cite{jia2021mbrs} introduces visible distortions, appearing as grid-like blocking artifacts on the stego image. These deviate from the image content, reducing imperceptibility and leaving detectable watermark traces. The FIN method~\cite{fang2023flow} exhibits low-frequency texture artifacts, evident in smooth regions with occasional white noise, yet offers better imperceptibility than grid artifacts of MBRS. In contrast, the proposed DRRW achieves superior visual quality, aligning more closely with human visual perception. Consequently, DRRW and DRRW-R attain enhanced imperceptibility, rendering watermark traces nearly imperceptible to subjective evaluation.
Quantitative evaluations further confirm the enhancement in visual quality. Figure~\ref{fig:realflow_psnr_ssim} summarizes the PSNR and SSIM results for each method. The proposed DRRW and DRRW-R methods achieve average PSNR values of \(40.9473\) dB and \(40.8971\) dB, respectively, with SSIM values of \(0.9995\) and \(0.9995\), indicating minimal embedding distortion. In contrast, the real-valued flow-based model R-DRRW yields a PSNR of \(33.9180\) dB and an SSIM of \(0.9971\). DRRW surpasses R-DRRW by \(7.03\) dB in PSNR and \(0.0024\) in SSIM. For the FIN method, the PSNR reaches \(39.7445\) dB and the SSIM \(0.9988\), with DRRW improving PSNR by \(1.2028\) dB and SSIM by \(0.0007\). The MBRS method records a PSNR of \(40.7302\) dB and an SSIM of \(0.9996\), where DRRW boosts PSNR by \(0.2171\) dB but slightly reduces SSIM by \(0.0001\). Although its quantitative metrics are close to those of DRRW, the presence of highly noticeable grid-like artifacts results in significantly lower subjective visual quality.  

In summary, DRRW and DRRW-R achieve the best visual quality in both subjective perception and objective evaluation metrics.

\begin{figure*}[t]
    \centering
    \begin{minipage}[]{.32\linewidth}
        \centering
        \includegraphics[width=\linewidth]{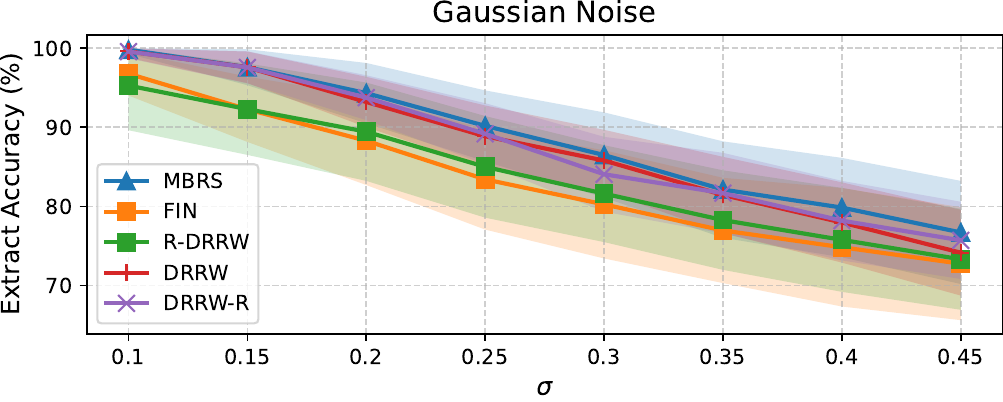}
    \end{minipage}
    \hspace{2pt}
    \begin{minipage}[]{.32\linewidth}
        \centering
        \includegraphics[width=\linewidth]{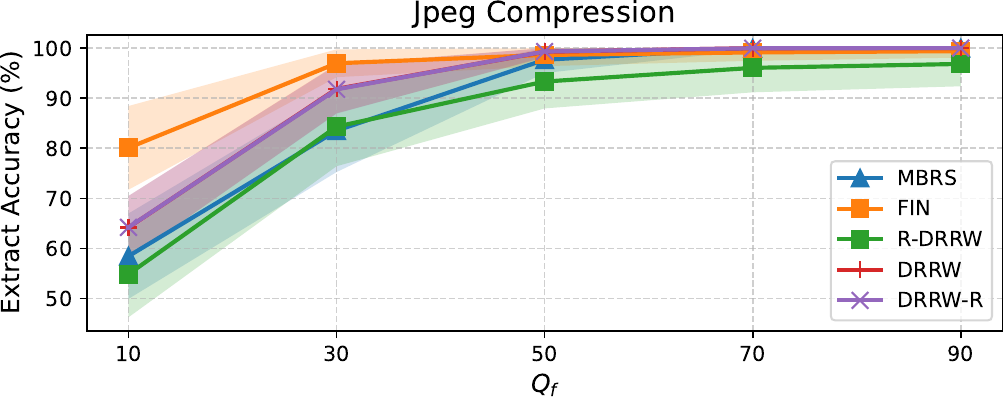}
    \end{minipage}
    \hspace{2pt}
    \begin{minipage}[]{.32\linewidth}
        \centering
        \includegraphics[width=\linewidth]{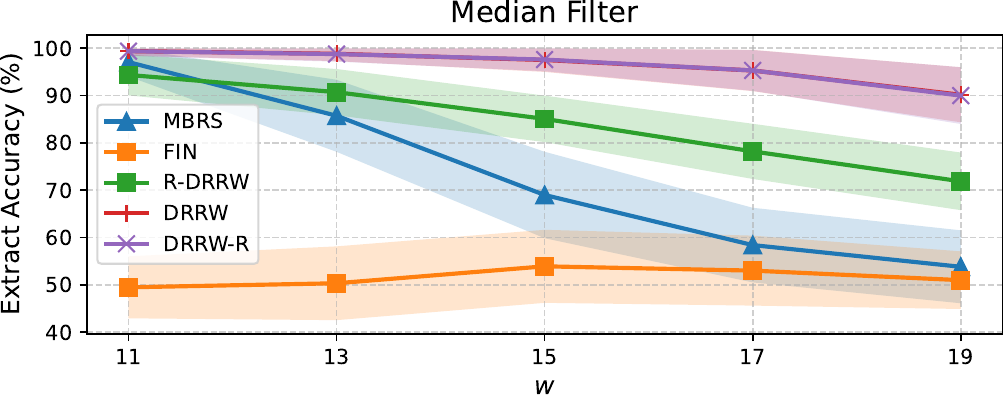}
    \end{minipage}
    \\
    \begin{minipage}[]{.32\linewidth}
        \centering
        \includegraphics[width=\linewidth]{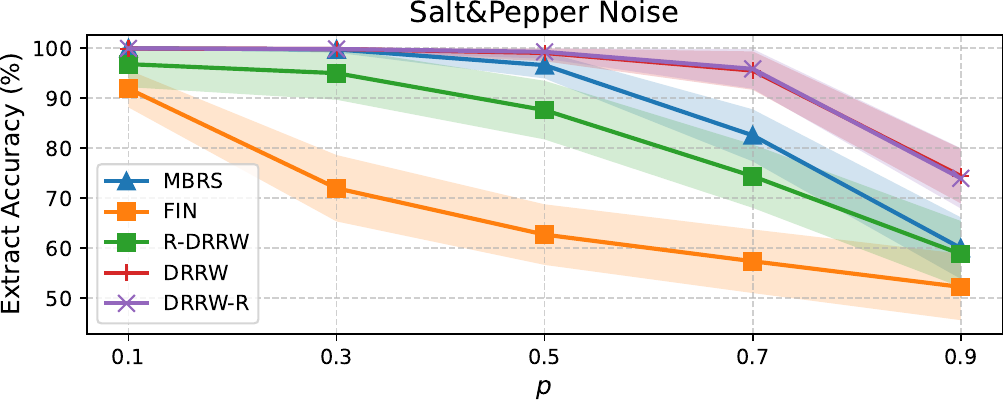}
    \end{minipage}
    \hspace{2pt}
    \begin{minipage}[]{.32\linewidth}
        \centering
        \includegraphics[width=\linewidth]{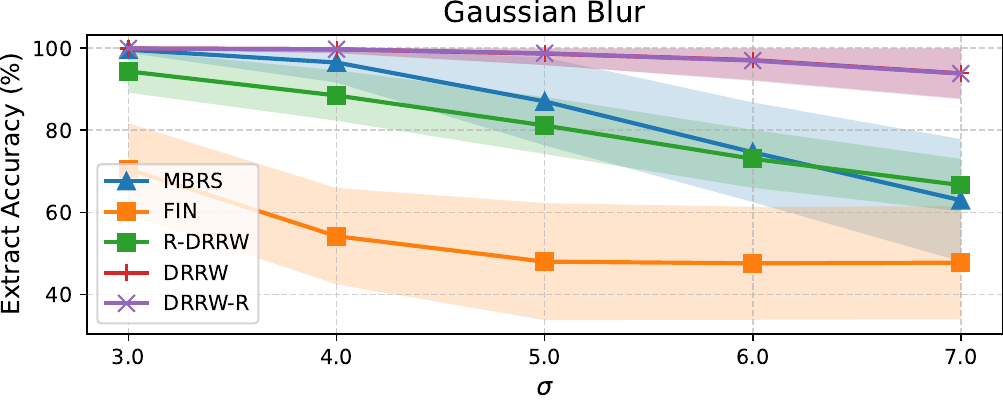}
    \end{minipage}
    \hspace{2pt}
    \begin{minipage}[]{.32\linewidth}
        \centering
        \includegraphics[width=\linewidth]{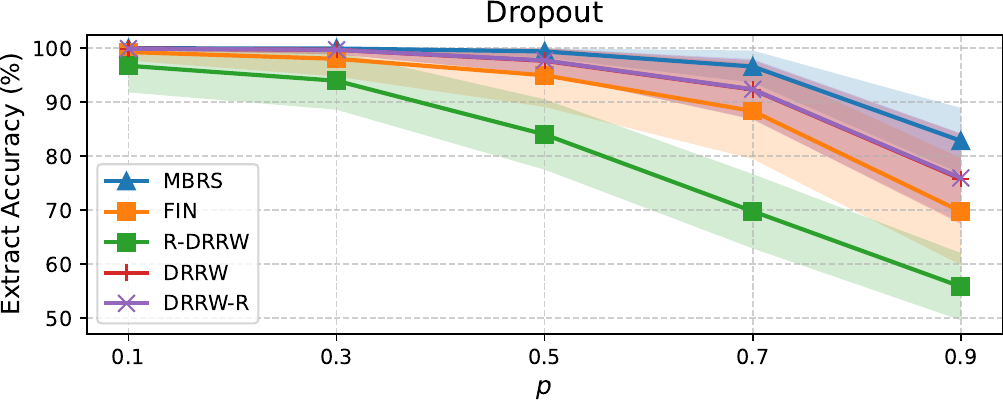}
    \end{minipage}
    \caption{Extraction accuracy (\%) comparison between MBRS~\cite{tang2024robust}, FIN~\cite{fang2023flow}, R-DRRW and the proposed DRRW under various distortions, including Dropout, Gaussian blur, median filtering, salt-and-pepper noise, Gaussian noise, and JPEG compression.}
    \label{fig:realflow_robustness_compare}
\end{figure*}
\subsubsection{Robustness Comparison}
We thoroughly assess the robustness of various methods against six prevalent digital distortions, illustrated in Figure~\ref{fig:realflow_robustness_compare}. The proposed integer flow-based DRRW and DRRW-R demonstrate overall superior robustness compared to the real-valued flow-based R-DRRW, MBRS~\cite{jia2021mbrs}, and FIN~\cite{fang2023flow} across most distortions.

Specifically, under median filtering (\(w=19\)), DRRW achieves \(90.01\%\), surpassing R-DRRW (\(85.12\%\)) by \(4.89\%\), MBRS (\(53.83\%\)) by \(36.18\%\), and FIN (\(51.01\%\)) by \(39.00\%\). For salt\&pepper noise (\(p=0.9\)), DRRW reaches \(74.46\%\), exceeding R-DRRW (\(65.28\%\)) by \(9.18\%\), MBRS (\(60.04\%\)) by \(14.42\%\), and FIN (\(52.27\%\)) by \(22.19\%\). Under Gaussian blur (\(\sigma=3.0\)), DRRW attains \(99.94\%\), outperforming R-DRRW (\(94.35\%\)) by \(5.59\%\), MBRS (\(99.65\%\)) by \(0.29\%\), and FIN (\(70.56\%\)) by \(29.38\%\). For Gaussian noise (\(\sigma=0.45\)), DRRW (\(74.14\%\)) aligns closely with MBRS (\(76.70\%\)) and other methods like DRRW-R (\(75.71\%\)) and R-DRRW (\(73.31\%\)), all outperforming FIN (\(72.75\%\)). Under dropout distortion, DRRW and DRRW-R achieve extraction accuracies of \(97.68\%\) and \(97.75\%\) at \(p=0.5\), slightly below MBRS (\(99.42\%\)), yet outperforming R-DRRW (\(84.03\%\)) by \(13.65\%\) and FIN (\(94.99\%\)) by \(2.69\%\). At \(p=0.9\), DRRW and DRRW-R reach \(75.81\%\) and \(75.98\%\), exceeding FIN (\(69.78\%\)) by \(6.03\%\) and MBRS (\(82.86\%\)) by \(6.95\%\), while R-DRRW (\(55.86\%\)) trails by \(19.95\%\) behind DRRW. For JPEG compression, DRRW and DRRW-R perform equally well, achieving \(64.18\%\) and \(64.21\%\) at \(Q_f=10\), surpassing MBRS (\(58.47\%\)) by \(5.71\%\) and R-DRRW (\(54.87\%\)) by \(9.31\%\), though FIN outperforms with \(80.09\%\). However, at \(Q_f \geq 30\), DRRW and DRRW-R approach \(100\%\), exceeding FIN (\(96.98\%\)) by about \(3\%\).

In summary, DRRW and DRRW-R outperform R-DRRW, MBRS~\cite{jia2021mbrs}, and FIN~\cite{fang2023flow} in robustness across most distortions, excelling in spatial-domain attacks (dropout, Gaussian blur, median filtering, salt-and-pepper noise) while competitive under JPEG compression and Gaussian noise. DRRW maintains robustness without reduction due to the second-stage reversible embedding. It should be noted that as a robust reversible watermarking method, DRRW uniquely recovers the cover image and watermark perfectly in a lossless channel, whereas other competitive methods cannot.

\begin{figure}[t!]
    \centering
    \begin{minipage}[]{\linewidth}
        \centering
        \includegraphics[width=\linewidth]{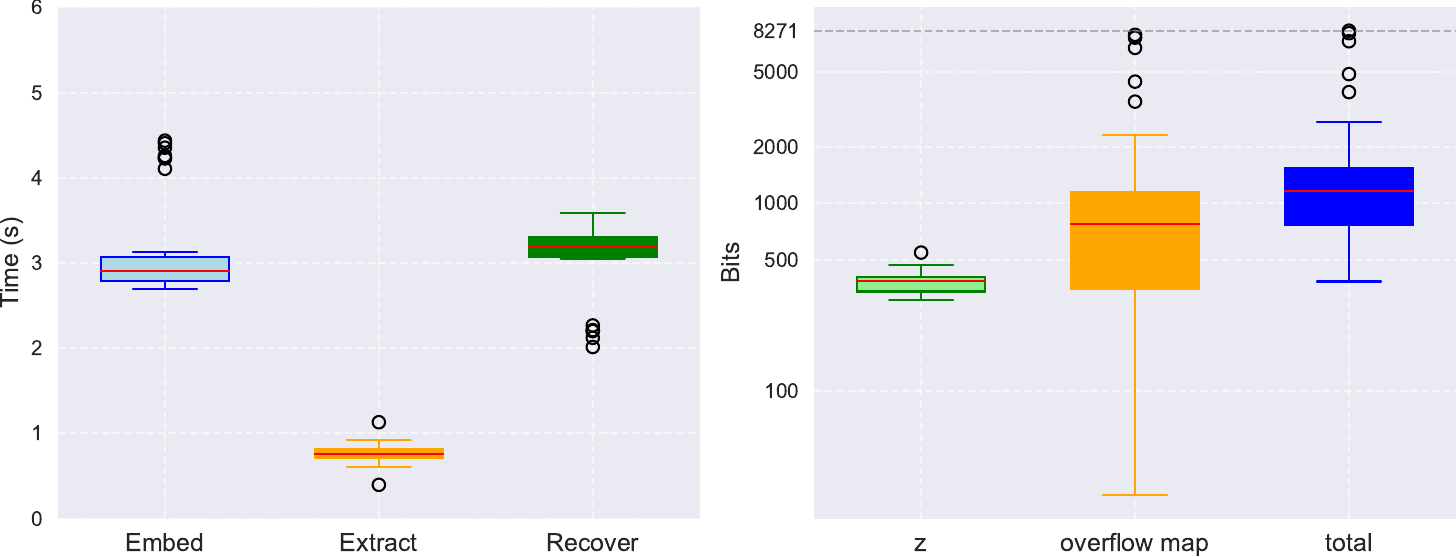}
    \end{minipage}
    \caption{The time complexity and auxiliary bitstream length of proposed DRRW trained on color images.}
    \label{fig:time_auxlens_drrw_color}
\end{figure}
\subsubsection{Auxiliary Bitstream Length and Time Complexity}
Figure~\ref{fig:time_auxlens_drrw_color} presents the auxiliary bitstream length and time complexity of the proposed DRRW trained on color images. Compared to PZM~\cite{tang2022highly} and STDM-PCET~\cite{tang2024robust} in Section~\ref{sec:compare_rrw}, DRRW reduces the auxiliary bitstream length to a range of \(384\) to \(8{,}271\) bits, with an average of \(1{,}962.19\) bits, achieving a \textbf{31.55$\times$} and \textbf{29.03$\times$} reduction over PZM and STDM-PCET, respectively.
More specifically, DRRW comprises an average of \(382.61\) bits for encoding the latent variable \(\mathbf{z}\) and \(1{,}558.58\) bits for the overflow map.
Furthermore, Figure~\ref{fig:time_auxlens_drrw_color} illustrates the time complexities of different methods. DRRW completes embedding in \(2.70\) to \(4.44\) seconds, averaging \(3.18\) seconds, significantly outperforming PZM, which requires \(195.17\) to \(286.40\) seconds (average \(248.65\) seconds), and STDM-PCET, which takes \(147.75\) to \(232.90\) seconds (average \(184.26\) seconds), achieving speedups of \textbf{78.19$\times$} and \textbf{57.94$\times$}, respectively. Extraction is completed in \(0.40\) to \(1.13\) seconds, averaging \(0.76\) seconds, compared to PZM’s \(6.86\) to \(10.01\) seconds (average \(8.54\) seconds) and STDM-PCET’s \(6.65\) to \(14.24\) seconds (average \(8.93\) seconds), resulting in \textbf{11.24$\times$} and \textbf{11.75$\times$} speedups. Recovery takes \(2.01\) to \(3.59\) seconds, averaging \(3.05\) seconds, outperforming PZM’s \(14.13\) to \(19.80\) seconds (average \(17.71\) seconds) and STDM-PCET’s \(12.70\) to \(20.12\) seconds (average \(15.89\) seconds), achieving \textbf{5.81$\times$} and \textbf{5.21$\times$} speedups.
Notably, DRRW successfully embedded watermarks in 16,762 images from PASCAL VOC 2012~\cite{pascal2012} without failures, addressing and improving the infeasibility of traditional RRW methods on large-scale datasets.
\vspace{-10pt}
\subsection{Ablation Experiments}\label{sec:ablation}  
This section conducts ablation experiments on three key components. First, we analyze how Penalty Loss affects auxiliary bitstream length. Second, we evaluate the influence of latent variable $\mathbf{z}$ loss on auxiliary bitstream length. Third, we examine the effectiveness of the training strategy in Section~\ref{sec:train_strategy}.

\subsubsection{Penalty Loss}  
To evaluate the impact of Penalty Loss in the DRRW framework, we conduct ablation experiments by adjusting the penalty weight $\lambda_p$ and analyzing its effects on auxiliary bitstream length. Specifically, $\lambda_p$ are set to $\{0, 10^2, 10^4, 10^6\}$, where $\lambda_p=0$ serves as the baseline without overflow penalty during training.

\begin{figure*}[t]
    \centering
    \begin{minipage}[]{.32\linewidth}
        \centering
        \includegraphics[width=\linewidth]{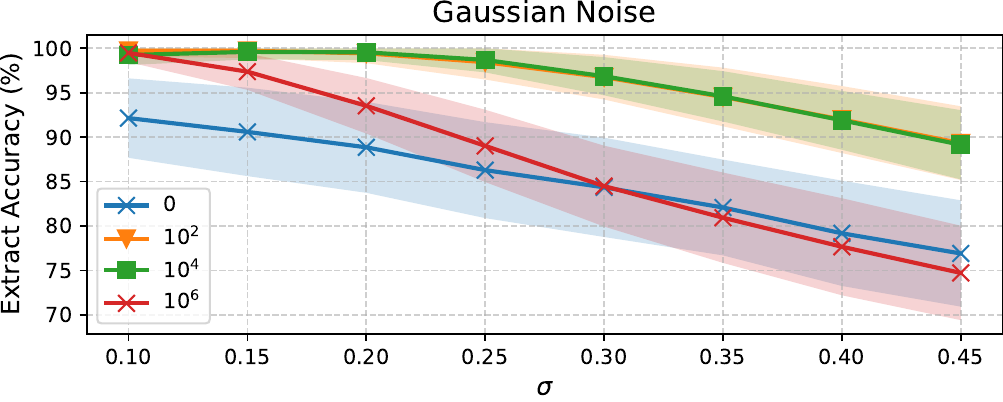}
    \end{minipage}
    \hspace{2pt}
    \begin{minipage}[]{.32\linewidth}
        \centering
        \includegraphics[width=\linewidth]{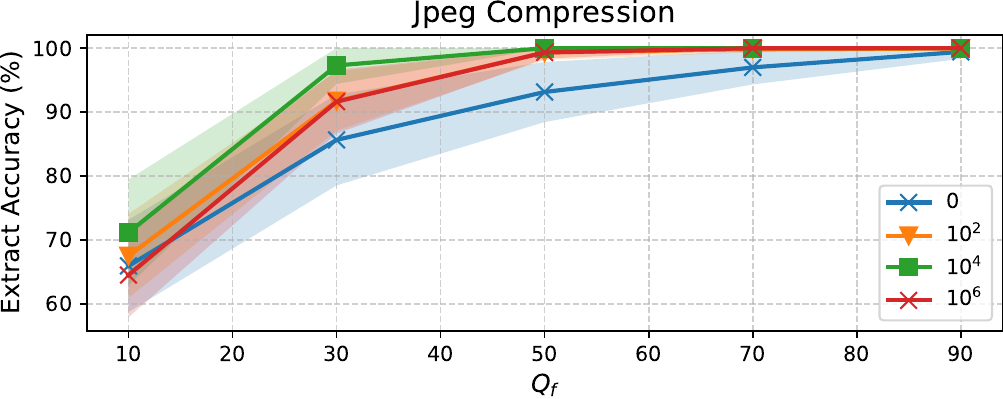}
    \end{minipage}
    \hspace{2pt}
    \begin{minipage}[]{.32\linewidth}
        \centering
        \includegraphics[width=\linewidth]{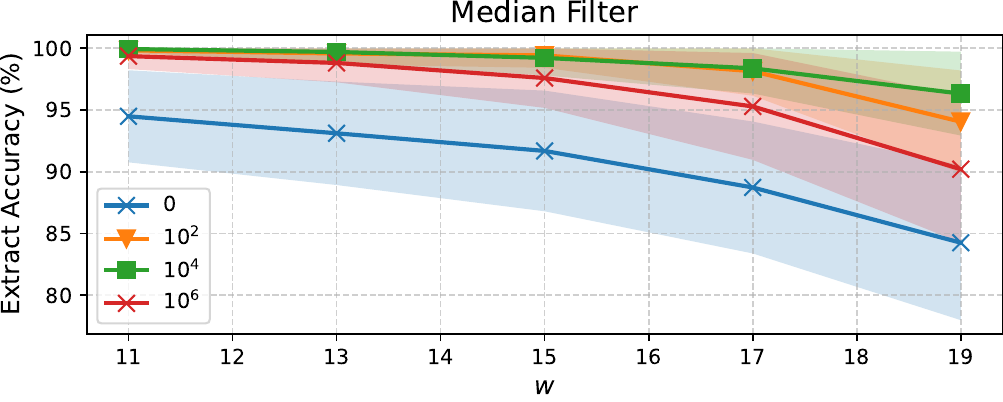}
    \end{minipage}
    \\
    \begin{minipage}[]{.32\linewidth}
        \centering
        \includegraphics[width=\linewidth]{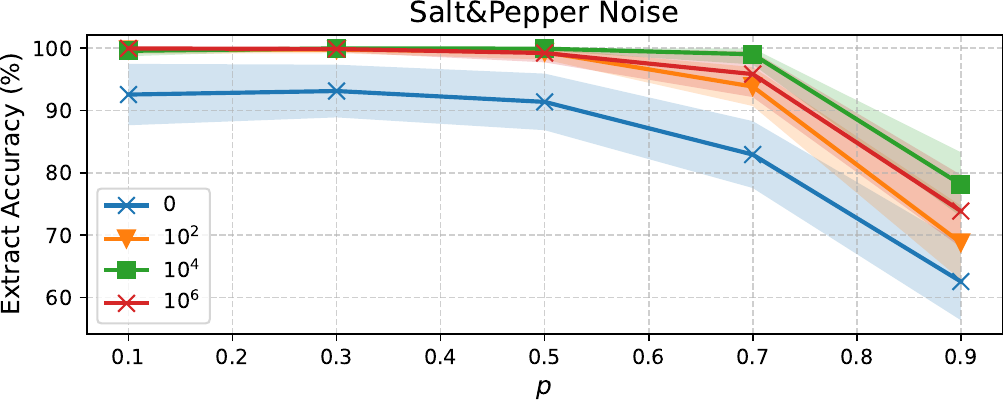}
    \end{minipage}
    \hspace{2pt}
    \begin{minipage}[]{.32\linewidth}
        \centering
        \includegraphics[width=\linewidth]{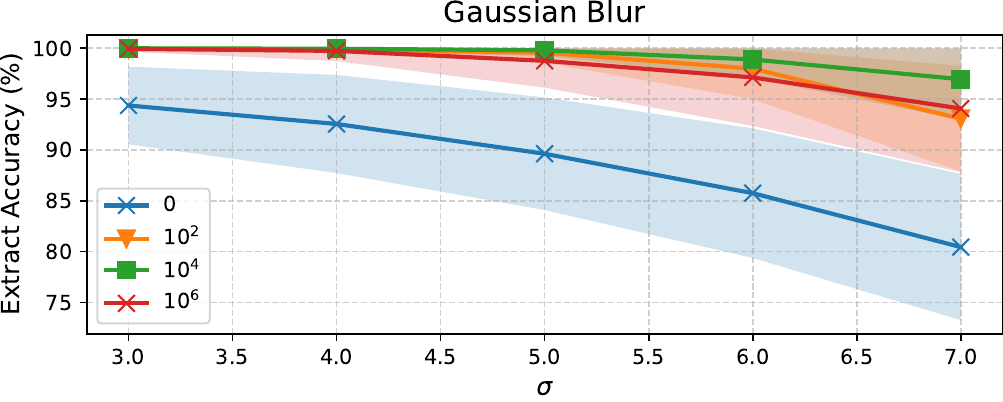}
    \end{minipage}
    \hspace{2pt}
    \begin{minipage}[]{.32\linewidth}
        \centering
        \includegraphics[width=\linewidth]{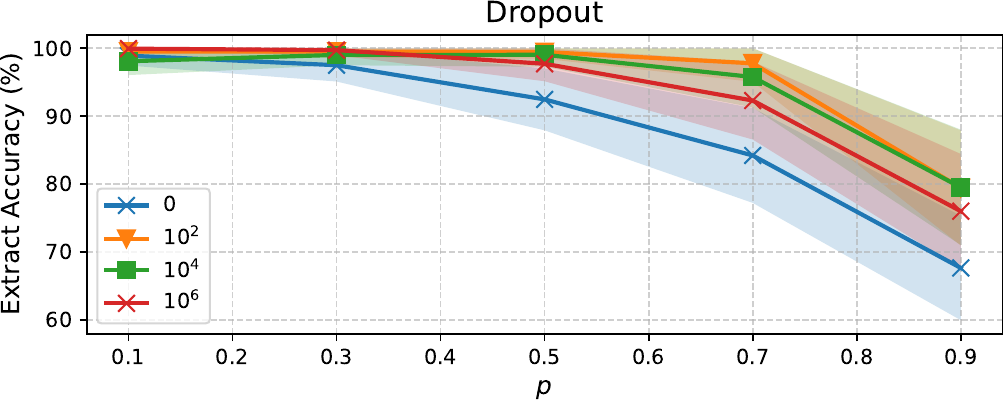}
    \end{minipage}
    \caption{DRRWs trained with different \(\lambda_p \in \{0, 10, 10^2, 10^4, 10^6\}\) exhibit varying extraction accuracies across six common distortions. The results show that Penalty Loss improves average extraction accuracy by over \(10\%\) under maximum distortion strengths.}
    \label{fig:penalty_compare}
\end{figure*}

\begin{figure}[t!]
    \centering
    \begin{minipage}[]{\linewidth}
        \centering
        \includegraphics[width=\linewidth]{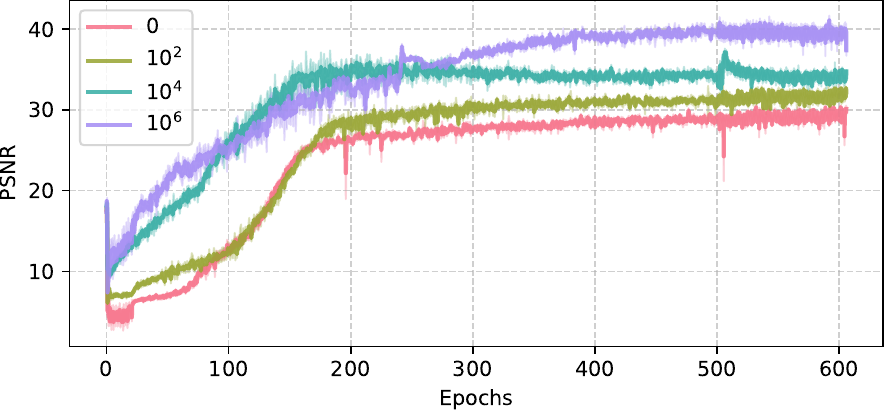}
    \end{minipage}
    \caption{PSNR curves of stego images over training epochs for different \(\lambda_p\).}
    \label{fig:penalty_psnr_loss}
\end{figure}

\begin{figure}[t]
    \centering
    \begin{minipage}[]{.46\linewidth}
        \centering
        \includegraphics[width=\linewidth]{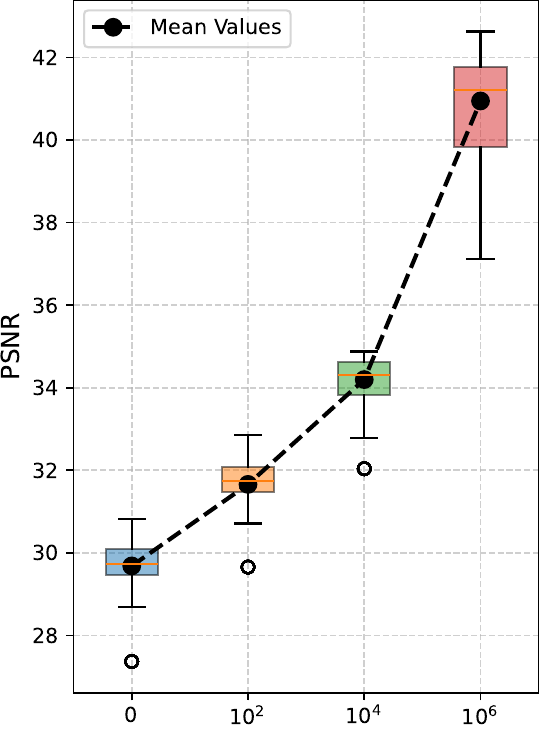}
    \end{minipage}
    \begin{minipage}[]{.5\linewidth}
        \centering
        \includegraphics[width=\linewidth]{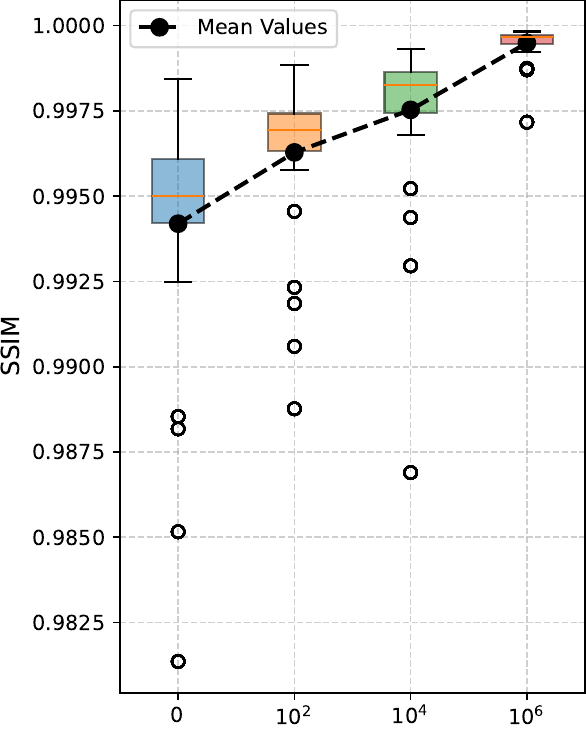}
    \end{minipage}
    \caption{PSNR and SSIM comparsion between DRRW under various \(\lambda_p\).}
    \label{fig:penalty_psnr_ssim}
\end{figure}
Before analyzing the impact of Penalty Loss on auxiliary bitstream length, we first assess its effect on the visual quality of the stego image. To better visualize the relationship between Penalty Loss weights \(\lambda_p\) and image quality, we plot PSNR curves over training epochs for various \(\lambda_p\) settings, as shown in Figure~\ref{fig:penalty_psnr_loss}. The results show that higher \(\lambda_p\) values consistently increase PSNR, demonstrating that Penalty Loss effectively enhances visual quality.
Furthermore, we report the average PSNR and SSIM of stego images after \(600\) training epochs for different \(\lambda_p\) values, as shown in Figure~\ref{fig:penalty_psnr_ssim}. With \(\lambda_p = 0\), the average PSNR and SSIM are \(29.6822\,\mathrm{dB}\) and \(0.9942\), respectively. Increasing \(\lambda_p\) to \(10^2\) raises these values to \(31.6585\,\mathrm{dB}\) and \(0.9963\), while \(\lambda_p = 10^4\) further improves them to \(34.1974\,\mathrm{dB}\) and \(0.9975\). At \(\lambda_p = 10^6\), the metrics reach \(40.9473\,\mathrm{dB}\) and \(0.9995\), reflecting gains of \(11.2651\,\mathrm{dB}\) in PSNR and \(0.0053\) in SSIM compared to \(\lambda_p = 0\) (\emph{i.e.}, no Penalty Loss). These results highlight that Penalty Loss significantly enhances watermark imperceptibility and improves stego image quality in the DRRW framework.
\begin{figure*}[t]
    \centering
    \begin{minipage}[]{.325\linewidth}
        \centering
        \includegraphics[width=\linewidth]{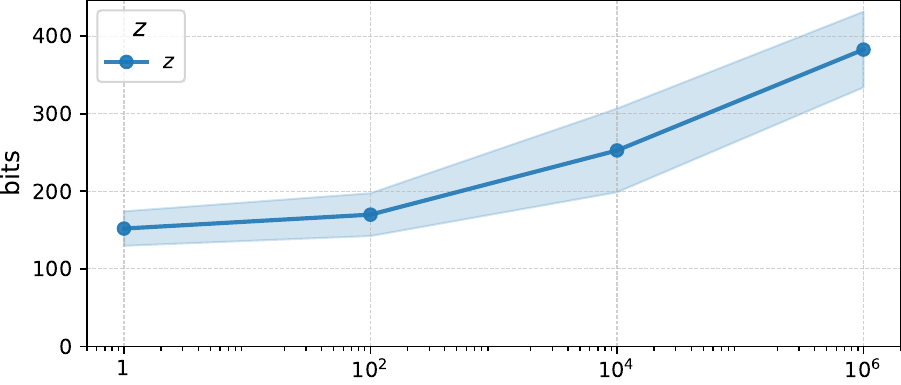}
    \end{minipage}
    \begin{minipage}[]{.325\linewidth}
        \centering
        \includegraphics[width=\linewidth]{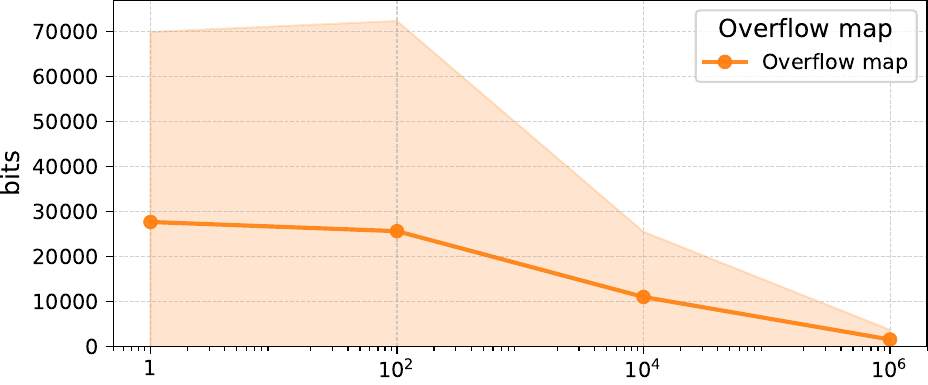}
    \end{minipage}
    \begin{minipage}[]{.325\linewidth}
        \centering
        \includegraphics[width=\linewidth]{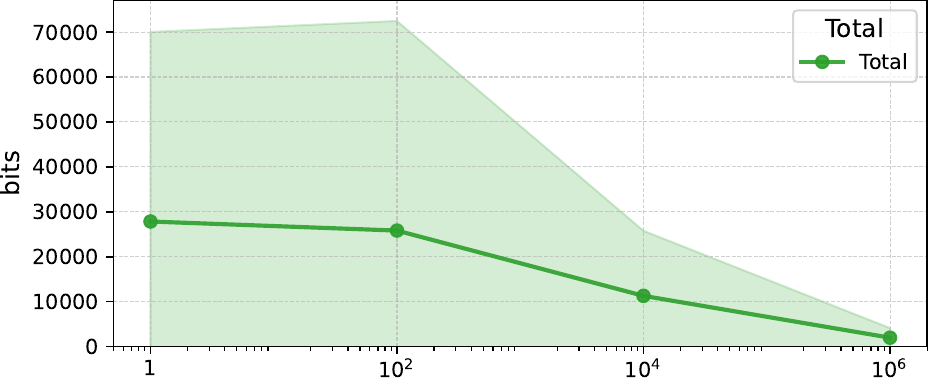}
    \end{minipage}
    \caption{Encoded bitstream lengths of the latent variable \(\mathbf{z}\) and the overflow map for different \(\lambda_p\) values. As \(\lambda_p\) increases, the bitstream length of \(\mathbf{z}\) gradually increases, while that of the overflow map decreases significantly.}
    \label{fig:penalty_auxlens}
\end{figure*}
We also evaluate the impact of Penalty Loss on watermark robustness under various distortion conditions with different \(\lambda_p\). Figure~\ref{fig:penalty_compare} shows that Penalty Loss significantly enhances robustness. Under six common distortions, it improves average extraction accuracy by over \(10\%\) at maximum distortion strengths. 
Finally, We analyze the impact of Penalty Loss on auxiliary bitstream length. Figure~\ref{fig:penalty_auxlens} shows that increasing \(\lambda_p\) has opposite effects on the bit lengths of the latent variable \(\mathbf{z}\) and the overflow map. As \(\lambda_p\) increases, the bit length for encoding \(\mathbf{z}\) gradually rises, while the overflow map bit length decreases significantly. This indicates that Penalty Loss reduces the number of overflow pixels in the first stage, thereby lowering the bits required for encoding the overflow map. Although the bit length of \(\mathbf{z}\) increases with \(\lambda_p\), it remains below \(400\) bits on average even at \(\lambda_p = 10^6\), causing negligible impact on reversible data hiding performance in the second stage.

In summary, Penalty Loss provides three key advantages within the DRRW framework: it significantly improves the visual quality of stego images, enhances watermark robustness, and effectively reduces the auxiliary bitstream length, greatly increasing the practicality of DRRW.

\begin{figure}[t]
    \centering
    \begin{minipage}[]{.52\linewidth}
        \centering
        \includegraphics[width=\linewidth]{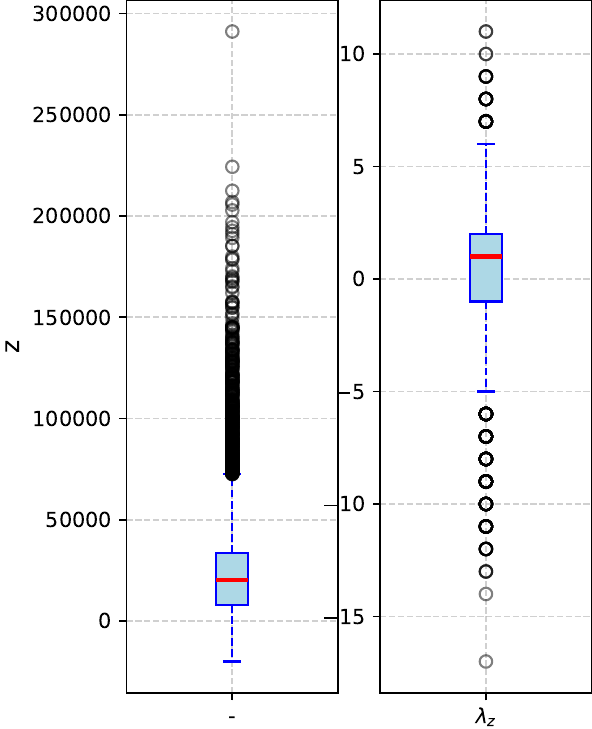}
    \end{minipage}
    \begin{minipage}[]{.46\linewidth}
        \centering
        \includegraphics[width=\linewidth]{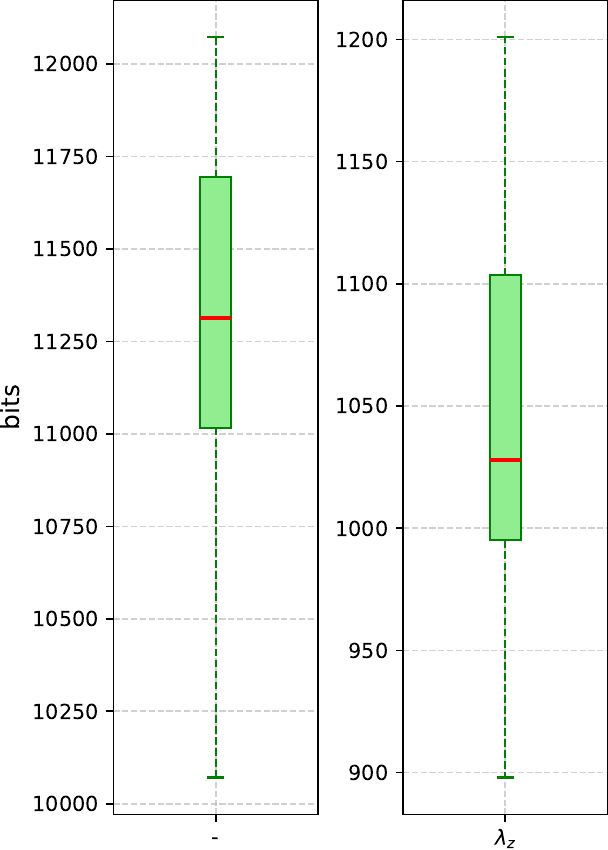}
    \end{minipage}
    \caption{Impact of regularization loss on the latent variable \(\mathbf{z}\) and its corresponding bitstream length (DRRW trained on grayscale images in Section~\ref{sec:compare_rrw}). The blue boxplot represents the case without regularization loss, while the green boxplot represents the case with regularization loss.}
    \label{fig:lambda_z}
\end{figure}
\subsubsection{Regularization loss for Latent Variable \(\mathbf{z}\)}\label{sec:ablation_lamda_z}
To justify the addition of regularization loss to the latent variable \(\mathbf{z}\), we observe that omitting this loss leads to a significant increase in the encoding bit length. Moreover, the network encounters numerical overflow issues during inference. To verify its necessity, we conduct the following ablation study.
As shown in Figure~\ref{fig:lambda_z}, without this loss function, the absolute values of \(\mathbf{z}\) can exceed \(3 \times 10^5\), mostly ranging from \(-5 \times 10^4\) to \(+5 \times 10^5\). This causes frequent computational overflow in both forward and inverse mappings, preventing complete image recovery. In the DRRW model for grayscale images (Section~\ref{sec:compare_rrw}), about \(46.123\%\) of test images fail to recover perfectly, while in the color model (Section~\ref{sec:compare_irw}), the failure rate is \(9.7\%\). As shown in Figure~\ref{fig:lambda_z},  After applying the regularization loss, \(\mathbf{z}\) is constrained within \([-18, +13]\), and the bitstream length is significantly reduced from \(11,292.83\) bits to \(1,046.03\), achieving a \(\textbf{10.79}\times\) reduction. This demonstrates that the regularization loss not only reduces the bitstream length of the latent variable \(\mathbf{z}\) but also solves the numerical overflow issue, as all test cover images are perfectly recovered without failure.

These results demonstrate that the regularization loss not only prevents numerical overflow but also improves encoding efficiency by reducing the bitstream length, ensuring reversibility. To ensure reliability, We recommend using \texttt{FLOAT64} precision during inference to avoid irreversibility caused by numerical precision errors.

\subsubsection{Training Strategy}  
\begin{figure}[t!]
    \centering
    \begin{minipage}[]{.95\linewidth}
        \centering
        \includegraphics[width=\linewidth]{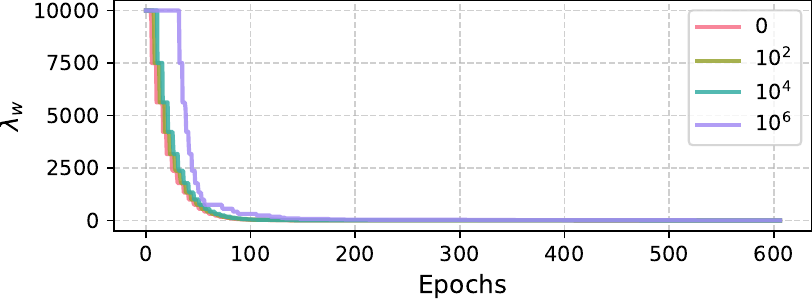}
    \end{minipage}
    \caption{Curves of \(\lambda_w\) over training epochs for different \(\lambda_p\), demonstrating its automatic adjustment and convergence.}
    \label{fig:train_lambda_secret}
\end{figure}

In Section~\ref{sec:train_strategy}, we introduce a training strategy that dynamically adjusts the watermark loss weight during training. To assess its effectiveness, we plot the curves of \(\lambda_w\) against training epochs for various \(\lambda_p\) settings, as shown in Figure~\ref{fig:train_lambda_secret}. The results reveal that \(\lambda_w\) stabilizes progressively as training advances. We further report the converged values of \(\lambda_w\) for \(\lambda_p \in \{0, 10^2, 10^4, 10^6\}\), which are \(0.312\), \(0.78\), \(2.53\), and \(7.53\), respectively. These findings show that the final \(\lambda_w\) increases with higher \(\lambda_p\). Remarkably, \(\lambda_w\) converges to a stable value irrespective of the \(\lambda_p\) choice. This proves that the proposed training strategy not only eliminates the need for manual adjustment of the watermark loss weight but also ensures convergence.

\section{Conclusion}\label{sec:conclusion}
This paper presented Deep Robust Reversible Watermarking (DRRW), a novel watermark method enabling lossless cover image recovery in lossless channels and robust watermark extraction in lossy channels. DRRW effectively overcomes the limitations of traditional robust reversible watermarking (RRW) methods, such as complex design, high time complexity, limited robustness, and impracticality for large-scale datasets. We developed an integer Invertible Watermark Network (iIWN) to map the cover-watermark pair and the stego image invertiblely. During training, DRRW utilizes an encoder-noise layer-decoder architecture to learn the robustness of various distortions. At inference, DRRW maps the cover-watermark pair into an overflowed stego image and latent variables, compresses them into a bitstream, and embeds it via reversible data hiding into the clipped stego image to achieve reversibility.

To improve practicality, DRRW introduced a Penalty loss, reducing the auxiliary bitstream length by \textbf{43.86$\times$} compared to existing RRW methods, while simultaneously improving the visual quality of stego image and watermark robustness, achieving a threefold benefit. Furthermore, we modeled an adaptive weight adjustment strategy that dynamically optimizes the watermark loss weight during training, removing manual tuning and ensuring stable convergence. Experiments demonstrate DRRW surpasses existing robust reversible and irreversible watermarking methods in robustness and visual quality, with computational efficiency improved by \textbf{55.14$\times$} for embedding, \textbf{5.95$\times$} for extraction, and \textbf{3.57$\times$} for recovery over existing RRWs. Remarkably, DRRW successfully achieves reversible embedding across all 16,762 images, significantly advancing the practical deployment of RRW.
\bibliographystyle{IEEEtran}
\bibliography{arxiv}
\vfill
\end{document}